%% file: 0_main.tex
\pdfoutput=1
\RequirePackage[svgnames]{xcolor}

\documentclass[11pt]{article}

\usepackage{acl2023}
\usepackage{amsmath}
\usepackage{amsmath,amssymb}
\usepackage{soul}
\usepackage{tikz}

\usetikzlibrary{shapes.geometric}
\usepackage{framed}
\usepackage{enumitem}
\newlist{RQ}{enumerate}{1}
\setlist[RQ]{label=\textbf{RQ\,\arabic*},ref={RQ\,\arabic*}}
\usepackage{comment}
\usepackage{natbib}
\usepackage{multibib}
\makeatletter
\usepackage{booktabs}
\usepackage[inkscapeformat=png]{svg}
\usepackage{graphicx}
\usepackage{caption}
\usepackage{subcaption}
\usepackage{tabularx}
\usepackage{soul}
\usepackage{float}
\usepackage{enumitem}
\usepackage{pifont}
\usepackage{arydshln}
\usepackage{lipsum}
\usepackage[T1]{fontenc}
\usepackage{pifont}
\usepackage{amsmath}
\usepackage{soul}
\usepackage[utf8]{inputenc}
\usepackage{inconsolata}
\usepackage{tikz}
\usepackage{arydshln}
\usepackage{lipsum}
\usepackage[normalem]{ulem}
\usepackage{wrapfig,graphicx,lipsum}
\usepackage{graphicx}
\usepackage{colortbl} 
\usepackage{listings} 

\usepackage[draft,textsize=footnotesize,textwidth=15mm]{todonotes}

\usepackage[many]{tcolorbox}

\tikzset{rndblock/.style={rounded corners,rectangle,draw,scale=0.8,outer sep=0pt}}

\newtcolorbox{defin}{colback=FireBrick!5!White,enhanced,title=Research Questions \& Contributions,
	attach boxed title to top left={xshift=0mm},boxrule=0pt,after skip=0cm,before skip=1cm,right skip=0cm,breakable,fonttitle=\bfseries,toprule=0pt,bottomrule=0pt,rightrule=0pt,leftrule=3pt,arc=0mm,skin=enhancedlast jigsaw,sharp corners,colframe=FireBrick!75!black,colbacktitle=FireBrick!65!black,
}

\usetikzlibrary{shapes.geometric, arrows}
\usetikzlibrary{decorations.markings}

\usepackage{fancybox}
\usepackage{xcolor}
\usepackage{hyperref}
 \definecolor{darkblue}{rgb}{0, 0, 0.5}
  \hypersetup{colorlinks=true, citecolor=darkblue, linkcolor=darkblue, urlcolor=darkblue}

\definecolor{vgreen}{HTML}{60A917}
\definecolor{vred}{HTML}{CE3A29}

\usepackage{xstring}
\usepackage{longtable}

\usepackage{tabularray}

\DefTblrTemplate{firsthead,middlehead,lasthead}{default}{
}
\DefTblrTemplate{firstfoot}{default}{
  \UseTblrTemplate{contfoot}{default}
  \UseTblrTemplate{caption}{default}
}
\DefTblrTemplate{middlefoot}{default}{
  \UseTblrTemplate{contfoot}{default}
  \UseTblrTemplate{capcont}{default}
}
\DefTblrTemplate{lastfoot}{default}{
  \UseTblrTemplate{note}{default}
  \UseTblrTemplate{remark}{default}
  \UseTblrTemplate{capcont}{default}
}

\newcolumntype{P}[1]{>{\centering\arraybackslash}p{#1}}

\usepackage{color}
\tcbuselibrary{skins}

\usepackage{setspace}
\usepackage[capitalise,nameinlink]{cleveref}

\crefname{section}{Sec.}{Sec.}

\usepackage{microtype}
\usepackage{hyperref}
\usepackage{graphicx}
\usepackage{comment}
\usepackage{amsmath}
\usepackage{amssymb}
\usepackage{algorithm}
\usepackage{algpseudocode}
\usepackage{colortbl}
\usepackage[export]{adjustbox} 
\usepackage{varwidth}
\usepackage{enumitem}
\setlist{leftmargin=1mm}
\usepackage{pifont}
\usepackage{booktabs}
\usepackage{multirow}
\usepackage{subcaption}
\usepackage{resizegather}
\usepackage{breqn}
\usepackage[capitalise]{cleveref}
\usepackage{graphicx}
\usepackage{tikz}
\usetikzlibrary{shapes.geometric, arrows}
\usetikzlibrary{decorations.markings}
\usepackage{soul}
\usepackage{wrapfig,graphicx,lipsum}
\usepackage{extsizes}
\usepackage{cuted}
\usepackage{flushend}
\usepackage{float}
\usepackage{changepage,threeparttable}
\usepackage{setspace}
\usepackage{caption}
\usepackage{booktabs}
\usepackage{dblfloatfix} 
\usepackage{fixltx2e}
\usepackage[normalem]{ulem}
\usepackage{lmodern}

\usepackage{environ}

\newlength{\myl}
\expandafter\let\expandafter\origequation\csname equation*\endcsname
\expandafter\let\expandafter\endorigequation\csname endequation*\endcsname
\long\def\[#1\]{\begin{equation*}#1\end{equation*}}
\RenewEnviron{equation*}{
  \settowidth{\myl}{$\displaystyle\BODY$} 
  \origequation
    \ifdim\myl>\linewidth
      \resizebox{\linewidth}{!}{$\displaystyle\BODY$}
    \else
      \BODY 
    \fi
  \endorigequation
}

\makeatletter
\newcommand{\DrawLine}{%
  \begin{tikzpicture}
  \path[use as bounding box] (0,0) -- (\linewidth,0);
  \draw[color=blue!75!black,dashed,dash phase=.5pt]
        (0-\kvtcb@leftlower-\kvtcb@boxsep,0)--
        (\linewidth+\kvtcb@rightlower+\kvtcb@boxsep,0);
  \end{tikzpicture}%
  }
\makeatother

\usepackage{pdfpages} 

\usepackage{times}
\usepackage{latexsym}

\usepackage[utf8]{inputenc}

\usepackage{microtype}

\usepackage{inconsolata}
\usepackage {soul}
\usepackage[symbol]{footmisc}

\DeclareRobustCommand{\augiefamily}{%
\fontfamily{augie}\fontseries{m}\fontshape{n} \selectfont}
\DeclareTextFontCommand{\textaugie}{\augiefamily}

\newcommand{\radiant}{\scalebox{0.8}[1.0]{\fontfamily{lmss}\selectfont \bfseries RADIANT }}

\newcommand{\ecd}{{\fontfamily{lmtt}\selectfont \bfseries ECD }}

\title{\textcolor{FireBrick}{\radiant}\hspace{-1.5mm}: \textcolor{FireBrick}{R}etrieval \textcolor{FireBrick}{A}ugmente\textcolor{FireBrick}{D} ent\textcolor{FireBrick}{I}ty-context \textcolor{FireBrick}{A}lig\textcolor{FireBrick}{N}men\textcolor{FireBrick}{T} - Introducing RAG-ability and Entity-Context Divergence}

\author{
 \textbf{Vipula Rawte\textsuperscript{1}\footnotemark[1]},
 \textbf{Rajarshi Roy\textsuperscript{1}},
 \textbf{Gurpreet Singh\textsuperscript{1}},
 \textbf{Danush Khanna\textsuperscript{1}},
 \textbf{Yaswanth Narsupalli\textsuperscript{1}},
 \\
 \textbf{Basab Ghosh\textsuperscript{1}},
 \textbf{Abhay Gupta\textsuperscript{1}},
 \textbf{Argha Kamal Samanta\textsuperscript{1}},
 \textbf{Abhay Gupta\textsuperscript{1}},
 \textbf{Aditya Shingote\textsuperscript{1}},\\
 \textbf{Aadi Krishna Vikram\textsuperscript{1}},
 \textbf{Vinija Jain\textsuperscript{2}\footnotemark[2]},
 \textbf{Aman Chadha\textsuperscript{3}\footnotemark[2]},
 \textbf{Amit Sheth\textsuperscript{1}},
 \textbf{Amitava Das\textsuperscript{1}}
\\
 \textsuperscript{1}AI Institute, University of South Carolina, 
\textsuperscript{2}Meta,
 \textsuperscript{3}Amazon GenAI 
}

\lstdefinestyle{listing1}{
    basicstyle=\scriptsize\ttfamily,  
    escapeinside={(*@}{@*)},          
    xleftmargin=-1em,                 
    framexleftmargin=-1em,             
    frame=single,                      
    columns=fullflexible,              
    aboveskip=0pt, belowskip=0pt,      
    showstringspaces=false,            
    keepspaces=true,                   
    framesep=0pt,                      
    numbersep=0pt                       
}

\begin{document}
\maketitle

\footnotetext[1]{Corresponding Author}
\footnotetext[2]{Worked independent of the position}

\begin{abstract}
As Large Language Models (LLMs) continue to advance, Retrieval-Augmented Generation (RAG) has emerged as a vital technique to enhance factual accuracy by integrating external knowledge into the generation process. However, LLMs often fail to faithfully integrate retrieved evidence into their generated responses, leading to factual inconsistencies. To quantify this gap, we introduce Entity-Context Divergence (\ecd \hspace{-2mm}), a metric that measures the extent to which retrieved information is accurately reflected in model outputs. We systematically evaluate contemporary LLMs on their ability to preserve factual consistency in retrieval-augmented settings, a capability we define as RAG-ability. Our empirical analysis reveals that RAG-ability remains low across most LLMs, highlighting significant challenges in entity retention and context fidelity. This paper introduces \radiant (Retrieval AugmenteD entIty-context AligNmenT), a novel framework that merges RAG with alignment designed to optimize the interplay between retrieved evidence and generated content. \radiant extends Direct Preference Optimization (DPO) to teach LLMs how to integrate provided additional information into subsequent generations. As a behavior correction mechanism, \radiant boosts RAG performance across varied retrieval scenarios, such as noisy web contexts, knowledge conflicts, and hallucination reduction. This enables more reliable, contextually grounded, and factually coherent content generation. Datasets are publicly available at: \url{https://huggingface.co/RADIANT-RAG}

\end{abstract}

\input{11_introduction}
\input{2_dataset}

\input{2_rag_ability}

\input{3_radiant}
\input{4_radiant_results}
\input{6_generalization}
\input{7_conclusion}
\input{8_limitations}

\bibliographystyle{acl_natbib}
\bibliography{custom}

\newpage
\input{10_appendix}

\end{document}

%% file: 11_introduction.tex
\section{Longer Context: No Assurance of Enhanced LLM Comprehension!}

LLMs have advanced textual processing by leveraging massive datasets and advanced architectures, yet they struggle with long-context inputs in tasks demanding comprehension and factuality. Although context windows now span thousands of tokens, effective use remains limited due to persistent biases, inefficiencies, and inconsistencies. Notably, simply expanding the context window does not guarantee improved performance, as inherent limitations remain. A key issue is the “lost in the middle” effect \cite{liu2023lost}, where models overemphasize text beginnings and ends while neglecting crucial mid-context content, undermining balanced information processing. Similarly, long in-context learning (LongICL) challenges models to maintain performance over extended token sequences, with benchmarks like LongICLBench revealing degradation in reasoning and semantic retention as complexity increases. Retrieval-augmented language models (RALMs) further complicate matters by merging internal knowledge with external evidence, often leading to conflicting information. Biases, including the Dunning-Kruger effect and confirmation bias \cite{jin2024knowledge}, cause over-reliance on familiar data, thereby compromising factual accuracy and limiting utility in precision-critical domains. These challenges underscore the need for innovative architectures and robust training strategies.

To overcome these challenges, we introduce \radiant (Retrieval AugmenteD entIty-context AligNmenT), a novel framework that seamlessly integrates retrieved evidence into generated responses. It employs metrics like Entity-Context Divergence (\ecd \hspace{-2mm}) to assess and enhance the alignment between entities and their contexts, ensuring consistent performance regardless of where critical information appears in the input.

\vspace{-7mm}
\begin{defin}

\begin{itemize}
[labelindent=-0.6em,labelsep=0.1cm,leftmargin=*]
\setlength\itemsep{0em}
\begin{spacing}{0.5}

\item[$\blacktriangleright$] 
{\footnotesize 
{\fontfamily{phv}\fontsize{8}{9}
\selectfont
\textbf{RQ1: To what extent is a given context comprehensible to an LLM, and how much of the provided information is effectively translated into the LLM's generation?}
\\ \\
We introduce \ul{RAG-ability}, a novel measure to quantify the comprehensibility and utilization of provided contexts in RAG tasks. RAG-ability evaluates how well an LLM integrates retrieved evidence into its output, bridging the gap between raw retrieval and coherent, contextually faithful generation. We introduce a novel metric called {Entity-Context Divergence (\ecd \hspace{-1.5mm})}.
}
}

\item[$\blacktriangleright$] 
{\footnotesize 
{\fontfamily{phv}\fontsize{8}{9}\selectfont
\textbf{RQ2: How can RAG-ability be improved?} \\ \\
We propose \ul{\radiant (Retrieval AugmenteD entIty-context AligNmenT), a paradigm that merges RAG with alignment principles}. \radiant optimizes the interplay between retrieved evidence and the model's internal representations, leveraging entity-context alignment techniques to enhance factuality, coherence, and utility in long-context reasoning tasks.}
}

\item[$\blacktriangleright$] 
{\footnotesize 
{\fontfamily{phv}\fontsize{8}{9}\selectfont
\textbf{RQ3: How sensitive is \radiant to context organization or information ordering?} \\ \\
Our framework systematically addresses mid-context biases, positional sensitivity, and the ``\emph{lost in the middle}'' phenomenon by ensuring robust alignment mechanisms that reduce the impact of context organization. \radiant demonstrates consistent performance regardless of the ordering or placement of critical information within the input.
}
}


\vspace{-3mm}
\end{spacing}
\end{itemize}

\end{defin}

%% file: 2_dataset.tex
\section{Related Work}

Retrieval-augmented generation (RAG) \cite{lewis2020retrieval} has emerged as a key area in generative AI, driven by advances in LLMs \cite{zhao2024retrieval,brown2020language,yang2024qwen2,dubey2024llama,jiang2023mistral,team2024gemma2b,openai2024gpt4technicalreport}. It tackles knowledge-intensive tasks by combining an LLM with a retriever that pulls relevant information from a document database \cite{li2023making,meng2024sfrembedding}. The retriever often uses embeddings \cite{zhao2024retrieval,meng2024sfrembedding} and re-ranking to gather context for the LLM to generate factually grounded answers. Recent work has introduced multi-step inference techniques to enhance RAG’s reliability further \cite{asai2023self,xu2023recomp,ke2024bridging}.

\section{Dataset and Experiment Setup}
This section outlines the selection of the LLM and the dataset preparation process.

\subsection{Choice of LLMs}

To better understand which types of LLM have comparative better or worse Rag-ability, we choose a wide array of LLMs. We deliberately selected both open-source and closed-source contemporary LLMs that have demonstrated outstanding performance across various NLP tasks. \\
\textbf{(A) Open-Source}: (i) LLAMA-3.8b \cite{dubey2024llama}, (ii) LLAMA-3.1-70b \cite{dubey2024llama}, (iii) Gemma-2-9b \cite{team2024gemma}. \\
\textbf{(B) Closed-Source}: (i) Gemini-1.5-pro \cite{team2024gemini}, (ii) Gemini-1.0-pro \cite{team2023gemini}, (iii) Qwen2-72B \cite{yang2024qwen2}).


\subsection{Prompts and RAG Context}

\paragraph{Prompts:} Our experiments primarily focus on the news domain, though we believe the findings are generalizable to other domains. For this study, we collected New York Times articles from 2023-2024, using the article titles as prompts in our experiments.

\paragraph{RAG Context:} To gather relevant information for a given query, we perform a Google search using the New York Times article abstract as the query and programmatically retrieve the top 15 documents. The text content from these websites is scraped and tokenized into sentences. Sentence embeddings are generated for each retrieved document using the Sentence-BERT model (all-MiniLM-L6-v2), and cosine similarity is calculated to rank sentences based on their relevance to the query. The top 30\% of ranked sentences across all documents are combined to construct the final context.



{\scriptsize
\begin{lstlisting}
(*@\textbf{Prompt}@*): Prime Minister of the United Kingdom visiting India.

(*@\textbf{Context}@*): As Rishi Sunak prepares to travel to India for the first time since he took office, UK officials believe the prime minister-born in Southampton to parents of Indian descent and with strong family ties there-will receive a warm welcome. 
...

Sunak is due to attend the G20 Summit in New Delhi on September 9 and 10.

(*@\textbf{Instruction}@*):
Generate a detailed precise and ordered article in proper format for the NY times tweet on the topic: {Prompt} and   
  context: {Context}
\end{lstlisting}
}

Our experiments employed structured prompts to guide AI models in generating contextually relevant and coherent content. The prompt's design is critical for ensuring precise and 
ordered responses.

%% file: 2_rag_ability.tex
\section{Entity-Context Divergence (\ecd \hspace{-2mm})}


Even when named entities (NEs) (e.g., people, places, organizations) are correctly mentioned in generated text, the context around these entities can shift, altering the meaning or factual accuracy of the content. Traditional evaluation metrics (such as cosine similarity) often focus on surface-level overlap or semantic similarity, which may fail to detect these subtle but impactful contextual changes. 

The \textbf{Entity-Context Divergence (\ecd \hspace{-2mm})} metric quantifies information drift in AI-generated content relative to a provided context, focusing on individual NEs - key carriers of information. NEs are anchor points for measuring such drift, as they remain unchanged across narratives while their surrounding context varies. Notably, two texts may contain the same set of NEs, yet if their contextual framing differs substantially, the narrative shift can be profound, resulting in a high information divergence. 

The following example demonstrates how the same NE, ``\emph{Rishi Sunak}'', appears in two distinct contexts - one in a political-economic setting discussing fiscal reforms and governance, and the other in a cultural-social setting emphasizing community engagement and artistic expression - highlighting information drift. 

{\scriptsize
\begin{lstlisting}
(*@\textbf{Text 1}@*): In a recent address to Parliament, UK Prime Minister (*@\textbf{Rishi}@*) (*@\textbf{Sunak}@*) outlined a sweeping economic recovery strategy that sought to address the multifaceted challenges posed by rising inflation, international trade disruptions, and post-pandemic uncertainty by proposing comprehensive fiscal reforms, targeted investments in green technology and critical infrastructure, and regulatory adjustments designed to foster sustainable growth and long-term fiscal stability.
    
(*@\textbf{Text 2}@*): At a bustling cultural festival in central London, UK Prime Minister (*@\textbf{Rishi}@*) (*@\textbf{Sunak}@*) engaged with local artists, community leaders, and residents in an open forum where he discussed the transformative power of creative expression, the importance of preserving historical heritage, and innovative ways to integrate artistic endeavors with urban renewal, thereby highlighting his commitment to enriching the nation's cultural landscape alongside its economic development.
\end{lstlisting}
}

\begin{figure*}[ht!]
    \centering
    \includegraphics[width=\textwidth]{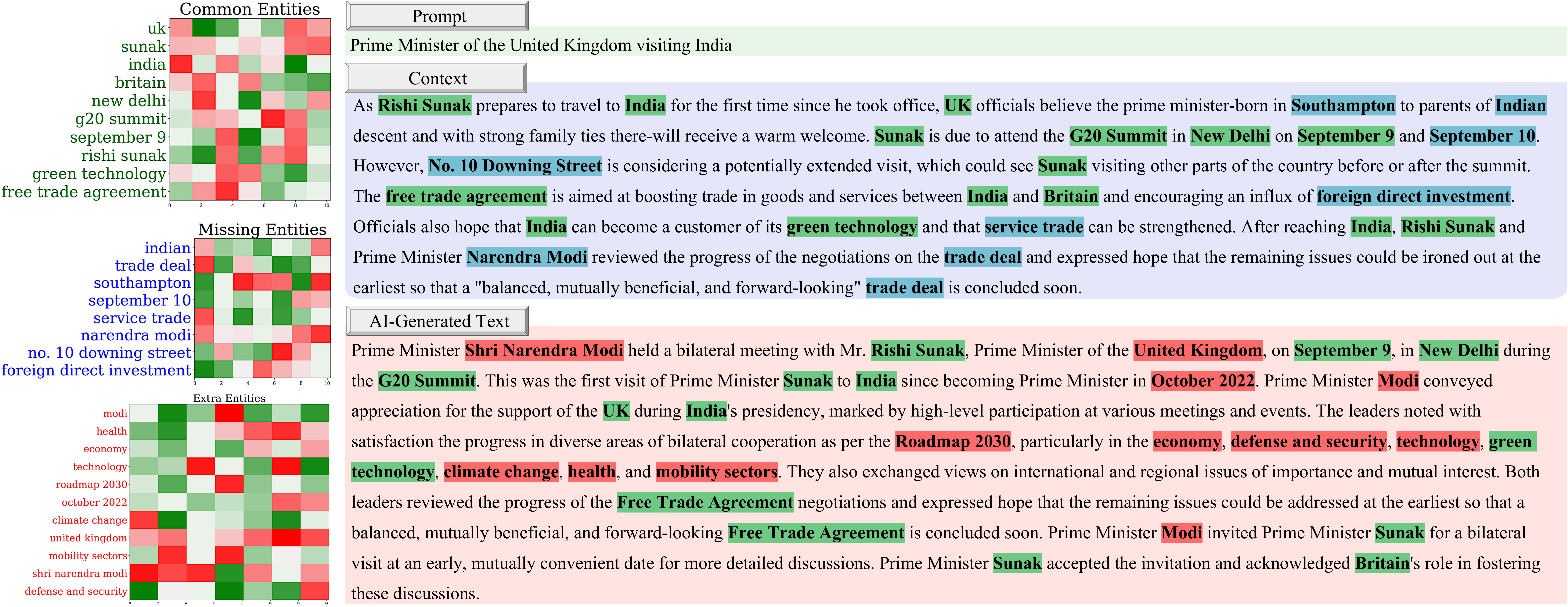}
    \caption{This figure illustrates Entity-Context Divergence (\ecd \hspace{-2mm}) by comparing the entities present in the provided context and the AI-generated text. The analysis categorizes entities into three groups: common, missing, and extra entities. The heatmaps (left) visually depict divergences for each entity and its context in the provided context vs. AI-generated content.}
    \label{fig:ECD_illustration}
    \vspace{-4mm}
\end{figure*}

By examining the surrounding words, (\ecd \hspace{-2mm}) identifies contextual shifts when named entities appear in different contexts. This interpretable framework consists of \textbf{three} key components:

\begin{itemize}
    \item \textbf{Common Entities:} Evaluates the divergence in contextual word distributions for common entities between the retrieved context and the generated output.

    \item \textbf{Missing Entity Penalty:} Penalizes instances where entities present in the provided context are omitted in the AI-generated content.

    \item \textbf{Added Entity Penalty:} Penalizes instances where the generated content introduces new (\emph{mostly hallucinated}) entities that do not appear in either the prompt or the provided context (\cref{fig:ECD_illustration}). 
\end{itemize}


\subsection{Divergence in Common Entities}
For each common entity $e \in E_{\text{common}}$, we calculate the contextual divergence using the Jaccard divergence of the surrounding context windows $W_r(e)$ and $W_g(e)$. The Jaccard divergence is defined as:
\[
    d_{\text{Jaccard}}(W_r(e), W_g(e)) = 1 - \frac{|W_r(e) \cap W_g(e)|}{|W_r(e) \cup W_g(e)|}
\]
where $|W_r(e) \cap W_g(e)|$ represents the number of overlapping words between the two windows, and $|W_r(e) \cup W_g(e)|$ represents the total number of unique words in both windows.

We use the following notations to define \ecd \hspace{-2mm}. Let $C_r$ represent the retrieved context and $C_g$ denote the AI-generated content. The sets of unique entities in $C_r$ and $C_g$ are represented as $E_r$ and $E_g$, respectively. For any entity, $e$, $W_r(e)$, and $W_g(e)$ refer to the sets of words within a window of size $w$ around $e$ in $C_r$ and $C_g$. The common entities between $C_r$ and $C_g$ are defined as $E_{\text{common}} = E_r \cap E_g$, with their count denoted by $n_{\text{common}} = |E_{\text{common}}|$. Finally, $\sigma$ represents the standard deviation of the \ecd \hspace{-2mm} distribution.

\subsection{Penalties for Missing Entities}
We define missing entities as those present in the provided context $C_r$ but absent in the AI-generated content $C_g$, represented as: $E_{\text{missing}} = E_r \setminus E_g$, where $\setminus$ denotes the set difference operator. The penalty for missing entities is calculated as:
\[
ME(C_r, C_g) = \frac{\sum_{e \in E_{\text{missing}}} \text{rank}(e) \cdot \sigma}{n_{\text{common}}}
\]
Here, $\text{rank}(e)$ is the rank of entity $e$ in the context, and $\sigma$ is the standard deviation of the \ecd \hspace{-2mm} distribution. 

\subsection{Penalties for Added Entities}
Added entities are those NEs which are present in the AI-generated content $C_g$ but absent in the given context $C_r$ and prompt, defined as: $E_{\text{added}} = E_g \setminus E_r$. The penalty for added entities is computed as:
\[
AE(C_r, C_g) = \frac{\sum_{e \in E_{\text{added}}} \text{rank}(e) \cdot \sigma}{n_{\text{common}}}
\]

where $\text{rank}(e)$ is the rank of entity $e$ in the AI-generated context, and $\sigma$ is the standard deviation of the \ecd \hspace{-2mm} distribution. 

\subsection{Calculating \ecd \hspace{-2mm}}
\begin{figure}[H]
    \centering
    \includegraphics[width=\columnwidth]{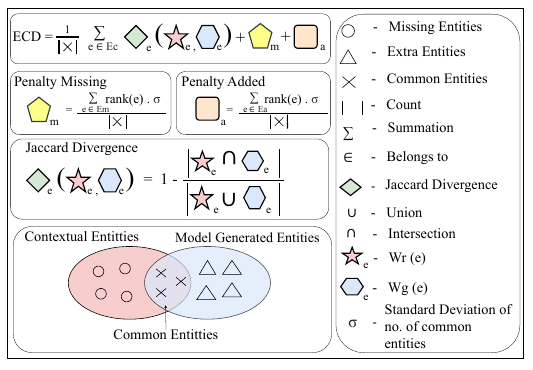}
    \caption{This visualization 
    captures the ability of \ecd \hspace{-2mm} to detect information drift by evaluating entity retention, omission, and addition in the AI-generated text.}
    \label{fig:ecd_equation_visual_illustration}
    \vspace{-4mm}
\end{figure}

\ecd \hspace{-2mm} combines the contextual divergence of common entities, the penalty for missing entities, and the penalty for the added entities. \cref{fig:ecd_equation_visual_illustration} visually illustrates our \ecd \hspace{-2mm} calculation process. The final \ecd \hspace{-2mm} score is computed as:
\begin{multline*}
    ECD(C_r, C_g) =\\ \frac{1}{n_{\text{common}}} \sum_{e \in E_{\text{common}}} \left( 1 - \frac{|W_r(e) \cap W_g(e)|}{|W_r(e) \cup W_g(e)|} \right)\\ + \frac{\sum_{e \in E_{\text{missing}}} \text{rank}(e) \cdot \sigma}{n_{\text{common}}}\\ + \frac{\sum_{e \in E_{\text{added}}} \text{rank}(e) \cdot \sigma}{n_{\text{common}}}
\end{multline*}

\vspace{-8mm}

\section{RAG-ability: Evaluating and Comparing LLMs' RAG Capabilities}

\begin{figure*}[]
    \centering
    \begin{minipage}{\textwidth}
        \begin{minipage}{0.49\textwidth}
            \centering
            \includegraphics[width=\textwidth]{images/1.pdf}
        \end{minipage}%
        \hfill
        \begin{minipage}{0.49\textwidth}
            \centering
            \includegraphics[width=\textwidth]{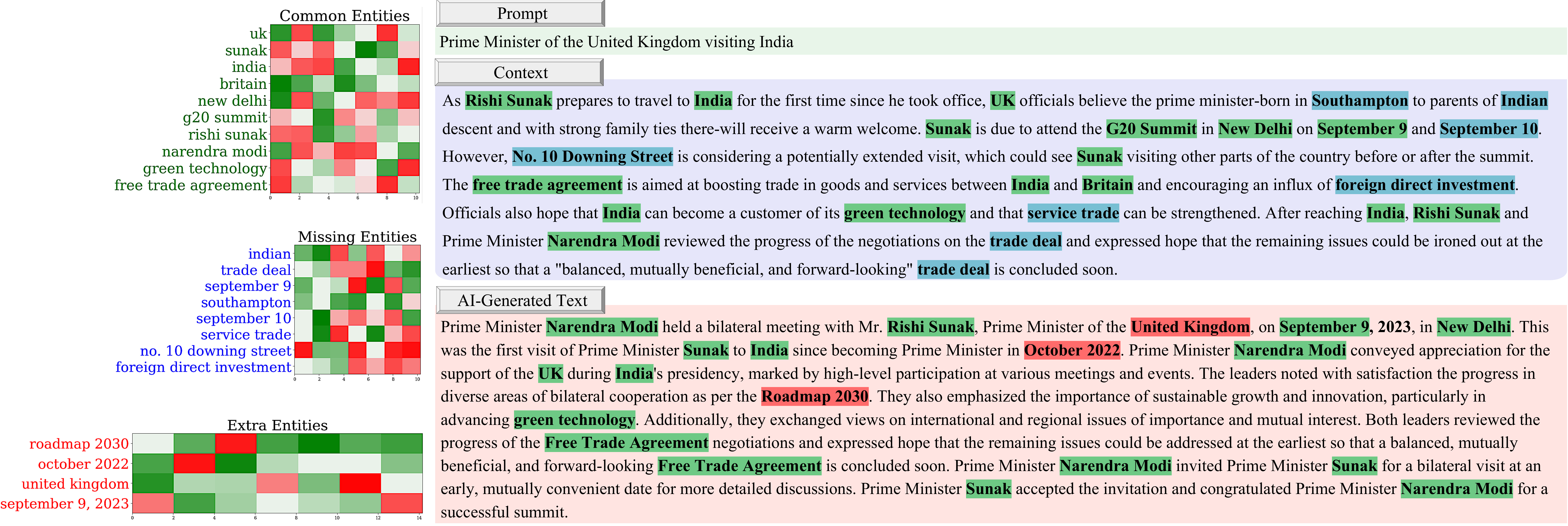}
        \end{minipage}\\[1em]
        \begin{minipage}{0.49\textwidth}
            \centering
            \includegraphics[width=\textwidth]{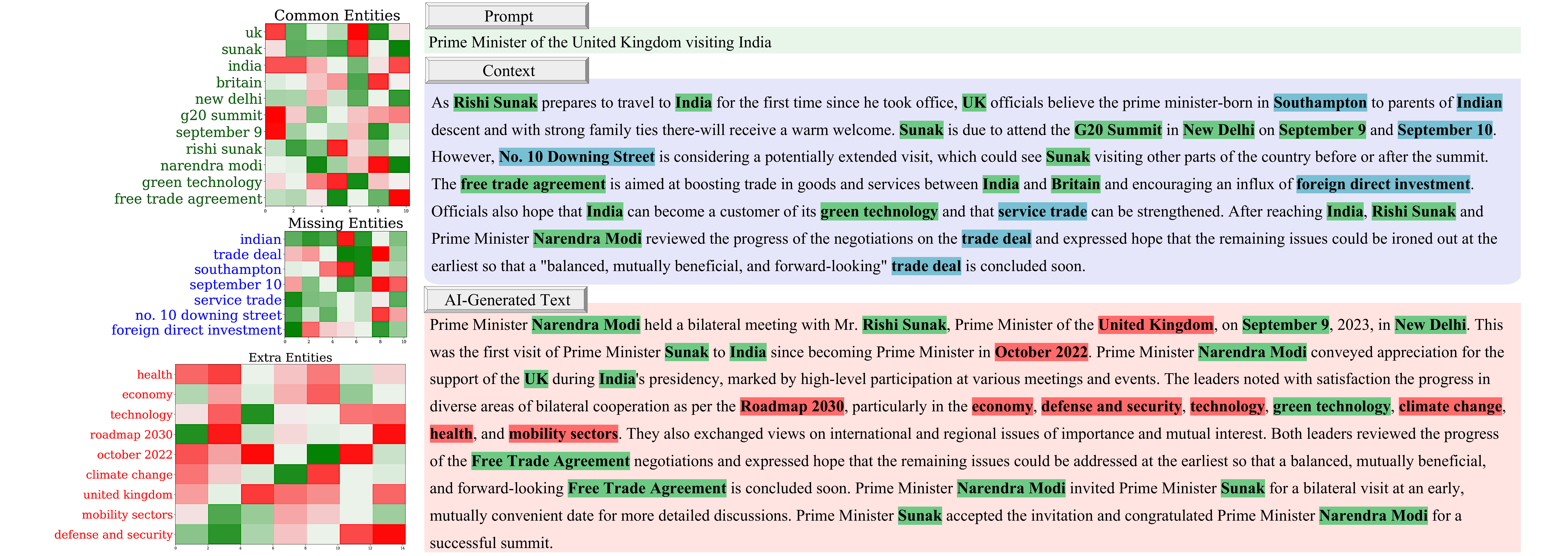}
            
        \end{minipage}%
        \hfill
        \begin{minipage}{0.49\textwidth}
            \centering
            \includegraphics[width=\textwidth]{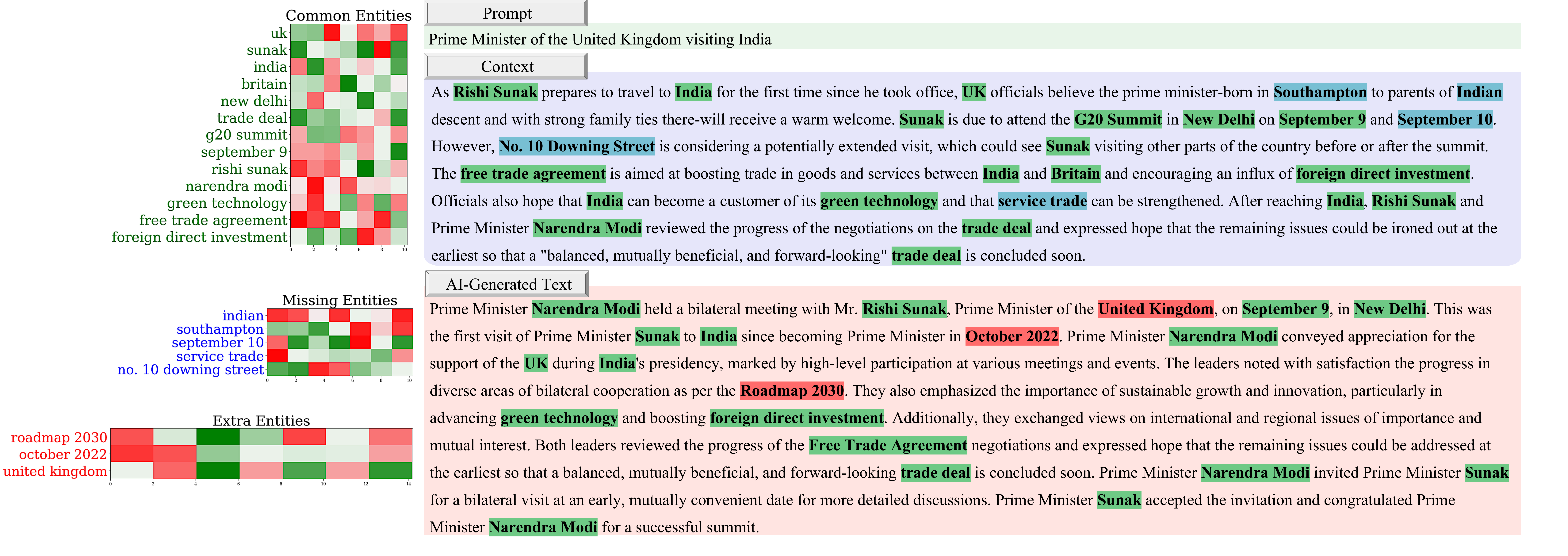}
        \end{minipage}
    \end{minipage}
    \centering
    \parbox{\textwidth}{\centering
    \caption{Combined RAG-Ability Results for AI Text Generation: (i) without any context, (ii) with perfect context (Twitter articles), (iii) with web context, and (iv) with synthesized context. Each subfigure shows (a) Extra Entities and (b) Missing Entities.}
    \label{fig:rag_ability_grid}}
\end{figure*}
\begin{figure*}[]
    \centering
    \begin{minipage}{0.85\textwidth} 
        \centering
        \begin{minipage}{0.49\textwidth}
            \centering
            \includegraphics[width=\textwidth]{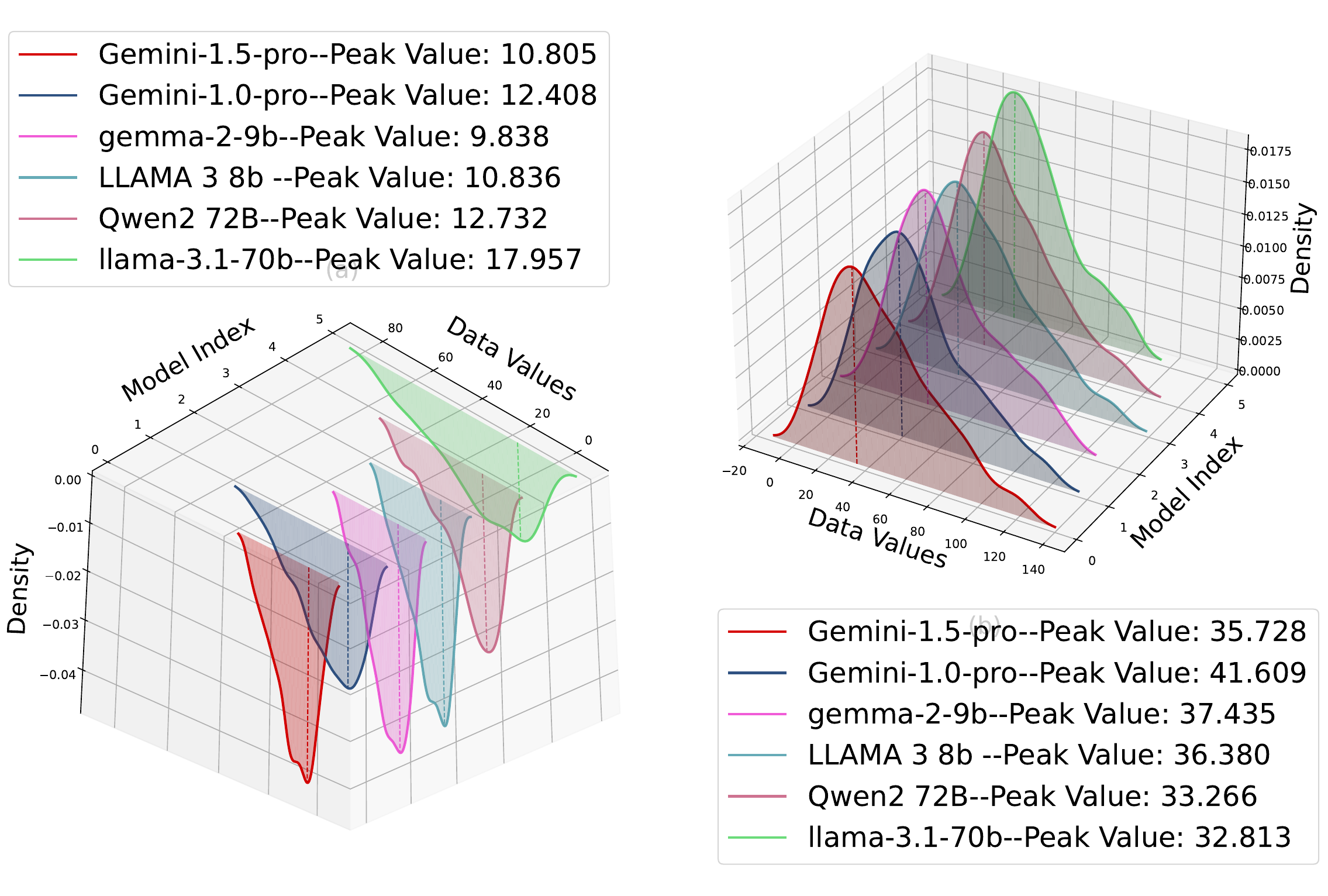}
        \end{minipage}%
        \hfill
        \begin{minipage}{0.49\textwidth}
            \centering
            \includegraphics[width=\textwidth]{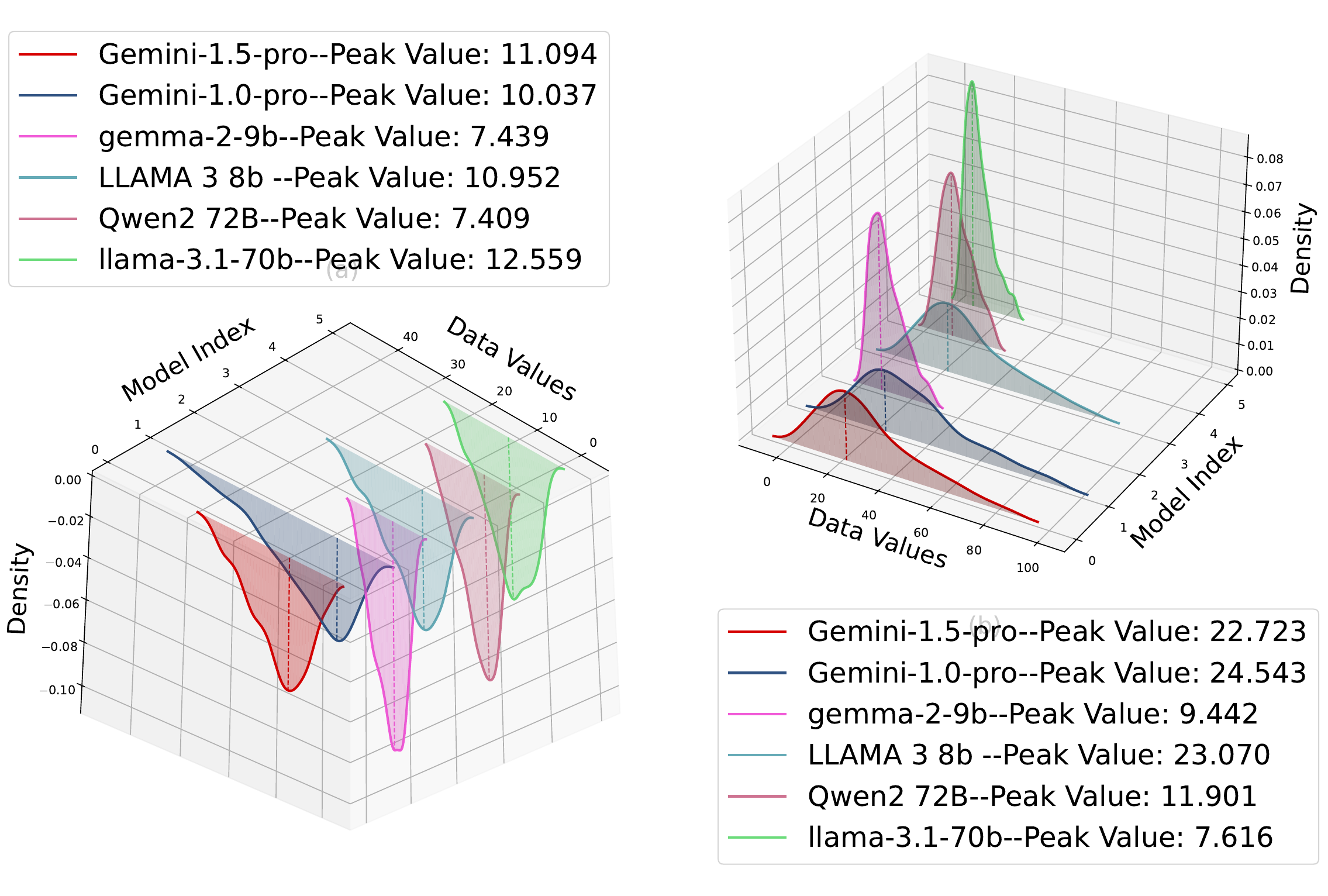}
        \end{minipage}\\[1em]
        
        \begin{minipage}{0.49\textwidth}
            \centering
            \includegraphics[width=\textwidth]{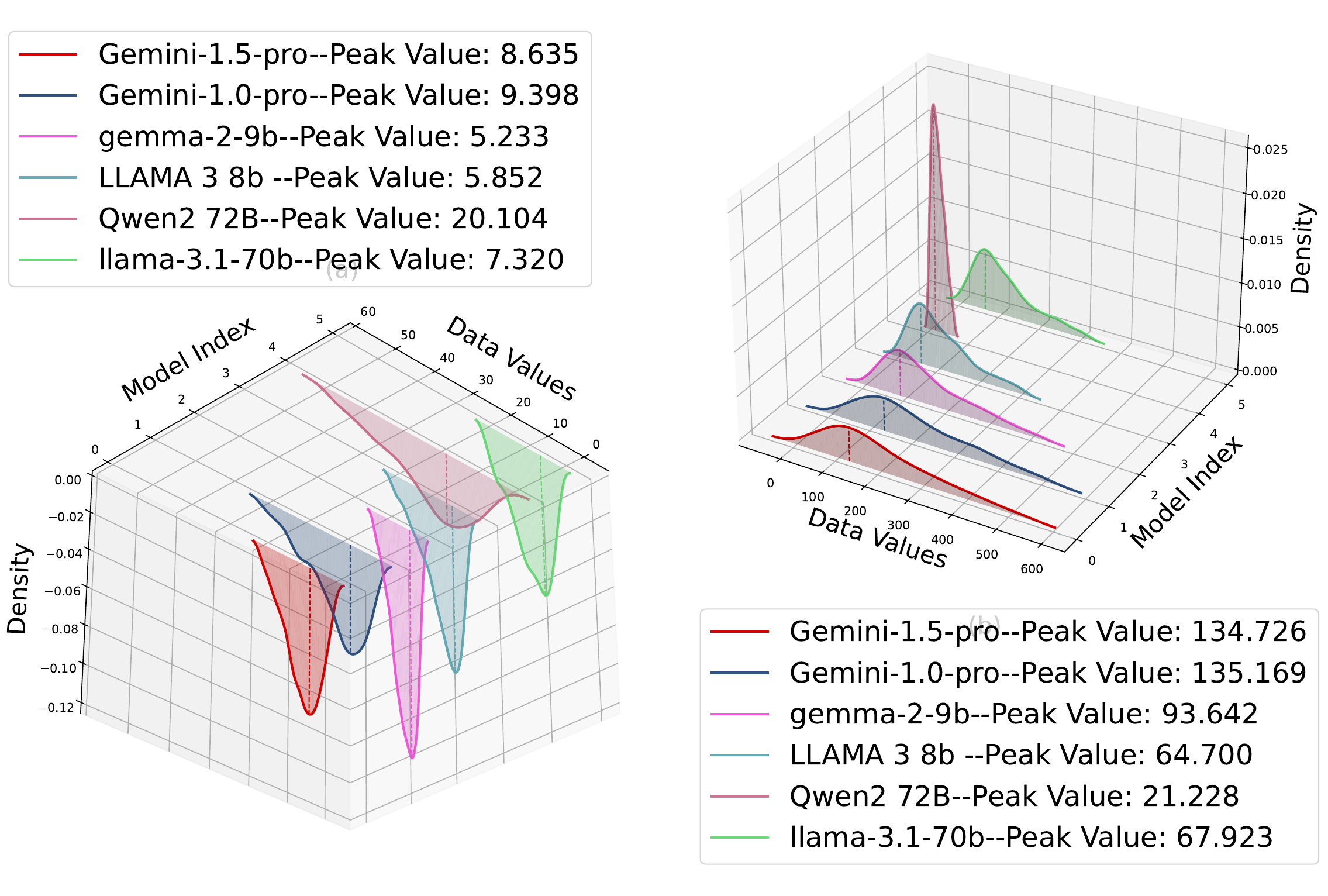}
        \end{minipage}%
        \hfill
        \begin{minipage}{0.49\textwidth}
            \centering
            \includegraphics[width=\textwidth]{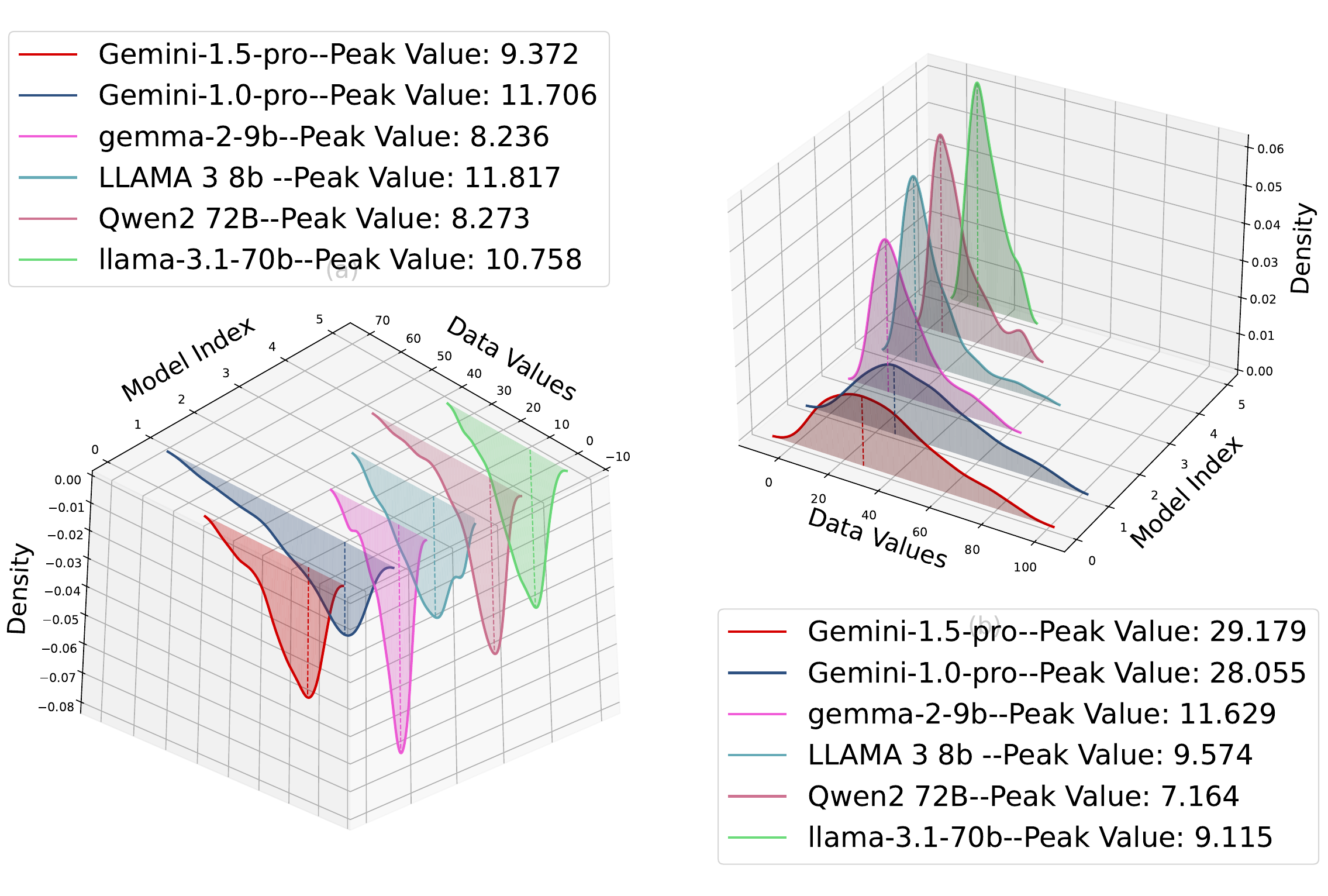}
        \end{minipage}
    \end{minipage}
    
    \vspace{1em}
    \parbox{0.9\textwidth}{\centering
        \caption{Combined RAG-Ability Results for AI Text Generation for \textbf{six} LLMs: (i) without any context, (ii) with perfect context (Twitter articles), (iii) with web context, and (iv) with synthesized context. Each subfigure shows the densities for: (a) Extra Entities and (b) Missing Entities.}
        \label{fig:rag_ability_grid_plots}
    }
\end{figure*}

We introduce \textbf{RAG-ability}, a robust evaluation framework designed to benchmark the RAG proficiency of LLMs. It offers a structured methodology to systematically measure how effectively LLMs integrate retrieved knowledge into their generated outputs. We present four scenarios in \cref{fig:rag_ability_grid,fig:rag_ability_grid_plots}: (a) without context, (b) with web-retrieved context, (c) with perfect context, and (d) with synthesized context. To demonstrate RAG-Ability’s effectiveness, we evaluate it across \textbf{six} LLMs (\cref{fig:rag_ability_grid_plots}).

\begin{figure}[H]
    \centering
    \includegraphics[width=\columnwidth]{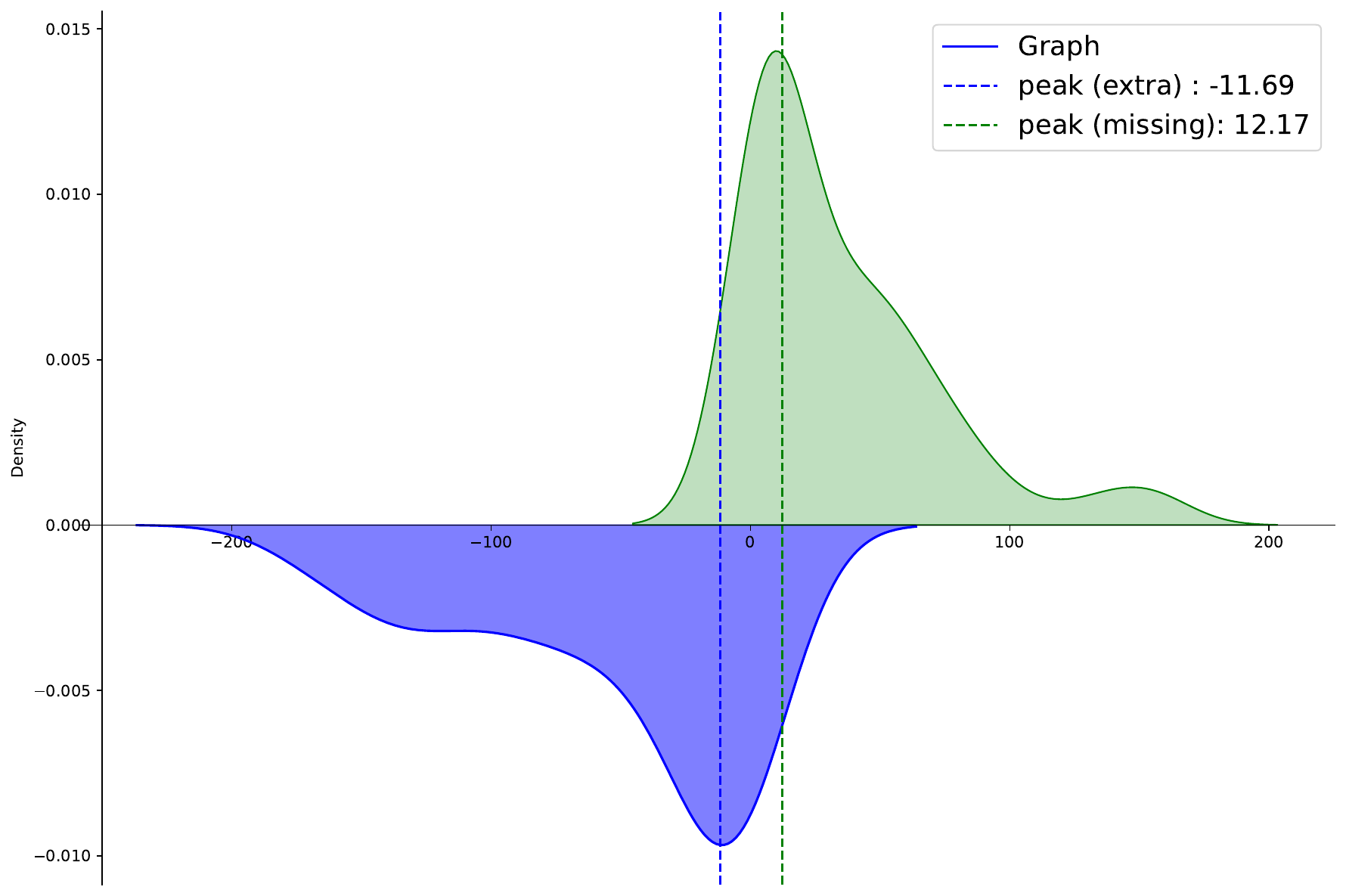}
    \caption{This plot visualizes the distribution of entity-context divergence (\ecd \hspace{-2mm}) by comparing missing entities (green region) and extra (hallucinated) entities (blue region) in AI-generated content. The dotted vertical lines denote the divergence peaks: extra entities at -11.69 (blue) and missing entities at 12.17 (green), highlighting the degree of entity-context drift in RAG.}
    \label{fig:ragability-extra-missing}
    \vspace{-4mm}
\end{figure}

Now, to illustrate behavioral characteristics of a given LLM's RAG-ability, let's imagine a density estimation plot as illustrated in the \cref{fig:ragability-extra-missing}. Green part depicts the common entity divergence+ missing entities ($d_{\text{Jaccard}}(W_r(e), W_g(e)) + ME(C_r, C_g)$) and the blue part corresponds the common entity divergence + added entities ($d_{\text{Jaccard}}(W_r(e), W_g(e)) + AE(C_r, C_g)$). The peak height of both distributions reflects the divergence scale of $(W_r(e), W_g(e))$. For each missing entity, the green plot shifts right by one \( \sigma_{\text{common-entities}} \), while for each added entity, the blue plot shifts left by the same amount. This visualization effectively captures how entity omissions and hallucinations impact the overall entity-context alignment, providing deeper insights into an LLM’s retrieval fidelity and generative consistency.

\begin{tcolorbox}[
left=9pt,right=2pt,colback=Navy!5!White,colframe=Navy!75!black,colbacktitle=Navy,
  title=\footnotesize \fontfamily{qbk} \selectfont \textbf{Takeaway from RAG-ability} ]
  
\vspace{-2mm}
\begin{itemize}
[labelindent=-0.2em,labelsep=0.1cm,leftmargin=0.3mm,rightmargin=0.3mm]
\setlength\itemsep{0.3em}
\begin{spacing}{0.9}

\scriptsize  \item[\ding{192}] 
{
{\fontfamily{phv}
\selectfont
\textbf{Lengthening along the Y-axis} indicates \textit{greater divergence}, which reflects a larger deviation or variability in the \ecd \hspace{-2mm} scores. This implies higher divergence in context and generated texts.
    
}
}
\vspace{-1mm} 
\scriptsize  \item[\ding{193}] 
{
{\fontfamily{phv}
\selectfont
\textbf{Lengthening along the X-axis} reflects a \textit{larger range of data}, implying a broader spread of \ecd \hspace{-2mm} scores or values captured in the analysis. This highlights the variability in the data range for extra and missing entities.

}
}
\vspace{-1mm} 
\item[\ding{118}] 
{\footnotesize 
{\fontfamily{phv}\fontsize{7}{8}
\selectfont

}
}

\vspace{-5mm}
\end{spacing}
\end{itemize}
\end{tcolorbox}

\clearpage  

%% file: 3_radiant.tex
\section{\radiant}

Preference alignment has become popular in calibrating LLMs for safer, more ethical outputs. Typically, it uses chosen and rejected response pairs. We use \ecd \hspace{-2mm} scores to distinguish preferred (low \ecd \hspace{-2mm}) from rejected (high \ecd \hspace{-2mm}) generations, raising the question: Can techniques like DPO minimize \ecd \hspace{-2mm} and enhance factual consistency?

The underlying intuition is that alignment training would encourage the model to read better, comprehend, and integrate contextual information, ultimately leading to more factually grounded and contextually aligned outputs.

\subsection{DPO-\ecd \hspace{-2mm} Objective}
The proposed DPO-\ecd \hspace{-2mm} (aka \radiant \hspace{-1mm}) objective integrates the DPO loss and \ecd \hspace{-2mm} alignment score into a single optimization objective. The goal is to maximize this objective concerning the policy $\pi$:

\vspace{-5mm}
\[
\adjustbox{max width=\columnwidth}{$
    \max_{\pi} \; \mathbb{E}_{(x, y^+, y^-)} \left[
    \underbrace{\log \frac{\pi(y^+ \mid x)}{\pi(y^- \mid x)}}_{\text{Statistical Preference Loss}}
    + \underbrace{\gamma \left( ECD(C_r, C_g^-) - ECD(C_r, C_g^+) \right)}_{\text{ECD Alignment Loss}}
    \right]
$}
\]

$ECD(C_r, C_g^+)$: The \ecd score for the preferred context $C_g^+$ relative to the retrieved context $C_r$. $ECD(C_r, C_g^-)$: The \ecd score for the non-preferred context $C_g^-$ relative to the retrieved context $C_r$. $\gamma$: A hyperparameter that controls the trade-off between preference-based loss and \ecd \hspace{-2mm} alignment.



%% file: 4_radiant_results.tex
\section{The Improved RAG-Ability}

\begin{figure}[h!]
    \centering
    \includegraphics[width=\columnwidth]{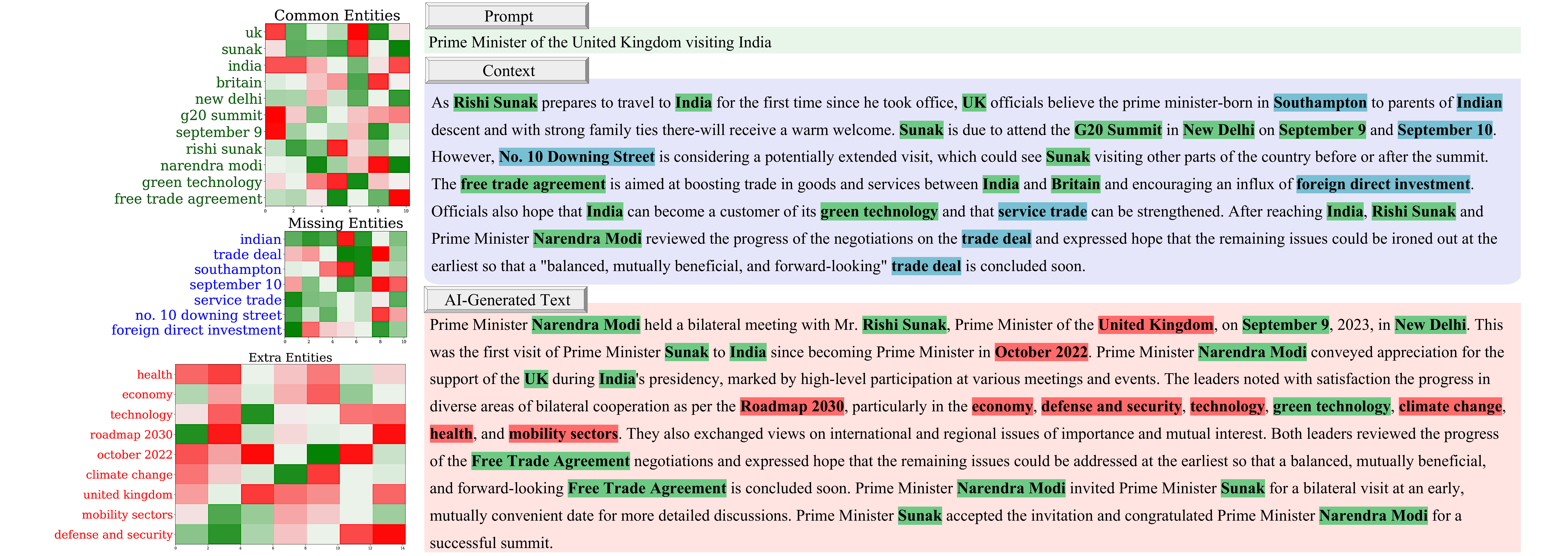}
    \caption{Illustration of the \radiant training objective in a RAG setup. The green and blue, zero-centered curves demonstrate significantly reduced peak magnitudes - indicating minimized \ecd \hspace{-2mm} - compared to the baseline behavior depicted in \cref{fig:ragability-extra-missing}.}
    \label{fig:radiant_goal}
    \vspace{-4mm}
\end{figure}

\begin{figure}[h!]
    \centering
    \includegraphics[width=\columnwidth]{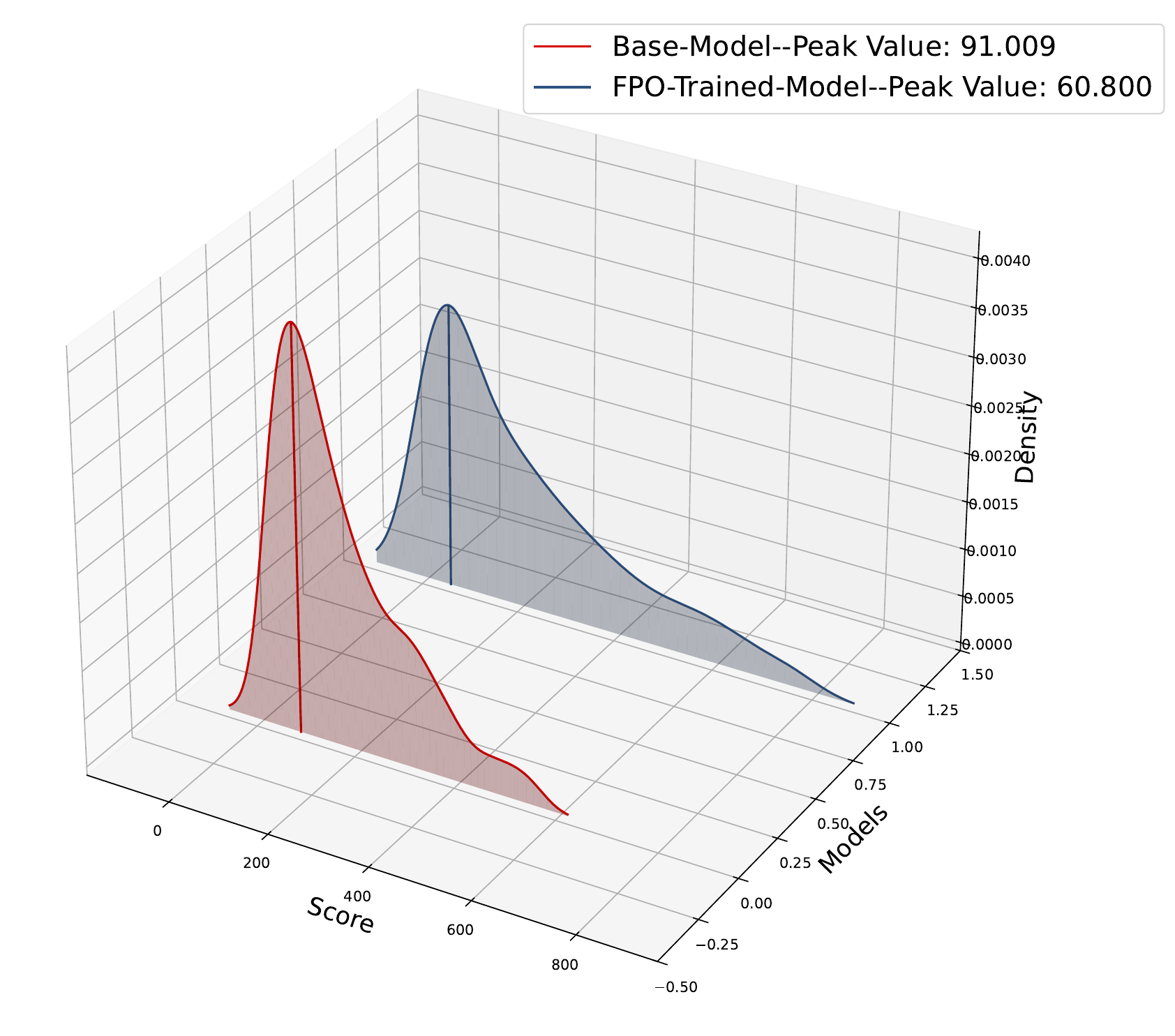}
    \caption{Comparison of \texttt{Gemma-7b-it} \ecd \hspace{-2mm} for Base and \radiant\hspace{-1mm}-trained (FPO).}
    \label{fig:Web-Retrieved_generation_scenario0}
\end{figure}

When training an LLM using the \radiant objective, the primary goal is to reduce the \ecd \hspace{-2mm} within a RAG setup. This objective is depicted in \cref{fig:radiant_goal}, where the green and blue curves are zero-centered and exhibit significantly lower peaks than those in \cref{fig:ragability-extra-missing}. Due to compute constraints, we only report improvements over one LLM \texttt{Gemma-7b-it} in the \cref{fig:Web-Retrieved_generation_scenario0}.

\section{Information Ordering}
A key issue is the ``lost in the middle'' effect \cite{liu2023lost}, where models overemphasize text beginnings and ends while neglecting critical mid-context content, leading to imbalanced information processing. This raises the question: Does the organization of context significantly impact AI generation (as measured by \ecd \hspace{-2mm})? After training LLMs with RADIANT, we provided 10 paraphrases of the same context. Empirical results consistently demonstrate that \radiant \hspace{-1.5mm}-trained LLMs overcome limitations imposed by variations in information organization.

\section{Which LLM Learns it Better?}

\begin{figure}[h!]
    \centering
    \includegraphics[width=\columnwidth]{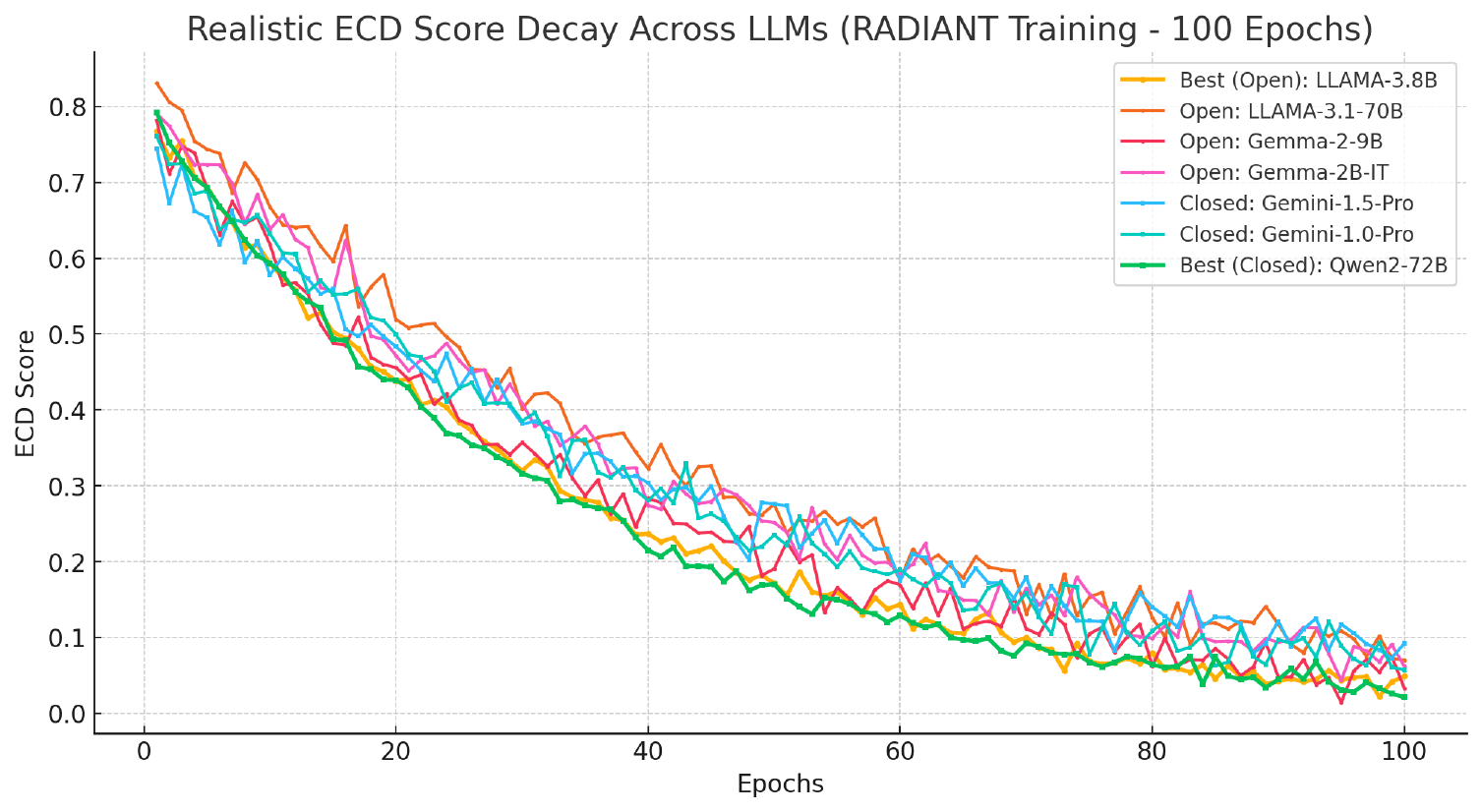}
    \caption{This figure shows how \ecd \hspace{-2mm} scores decrease over the epochs during \radiant training.}    
    \label{fig:which_llm_learns}
\end{figure}

LLMs trained with \radiant exhibit decreasing \ecd \hspace{-2mm} scores over training epochs. Among the sourced models, \texttt{LLaMA-3.8B} demonstrates the best calibration, followed by \texttt{LLaMA-3.1-70B}. \texttt{Gemma-2-9B} performs slightly worse, highlighting the effectiveness of \radiant in enhancing model stability and confidence alignment (\cref{fig:which_llm_learns}).

%% file: 6_generalization.tex
\section{Assessing Generalization \& Model Quality}  
\label{sec:auto-hvi}

\begin{figure}[h!]
    \centering
    \includegraphics[width=\columnwidth]{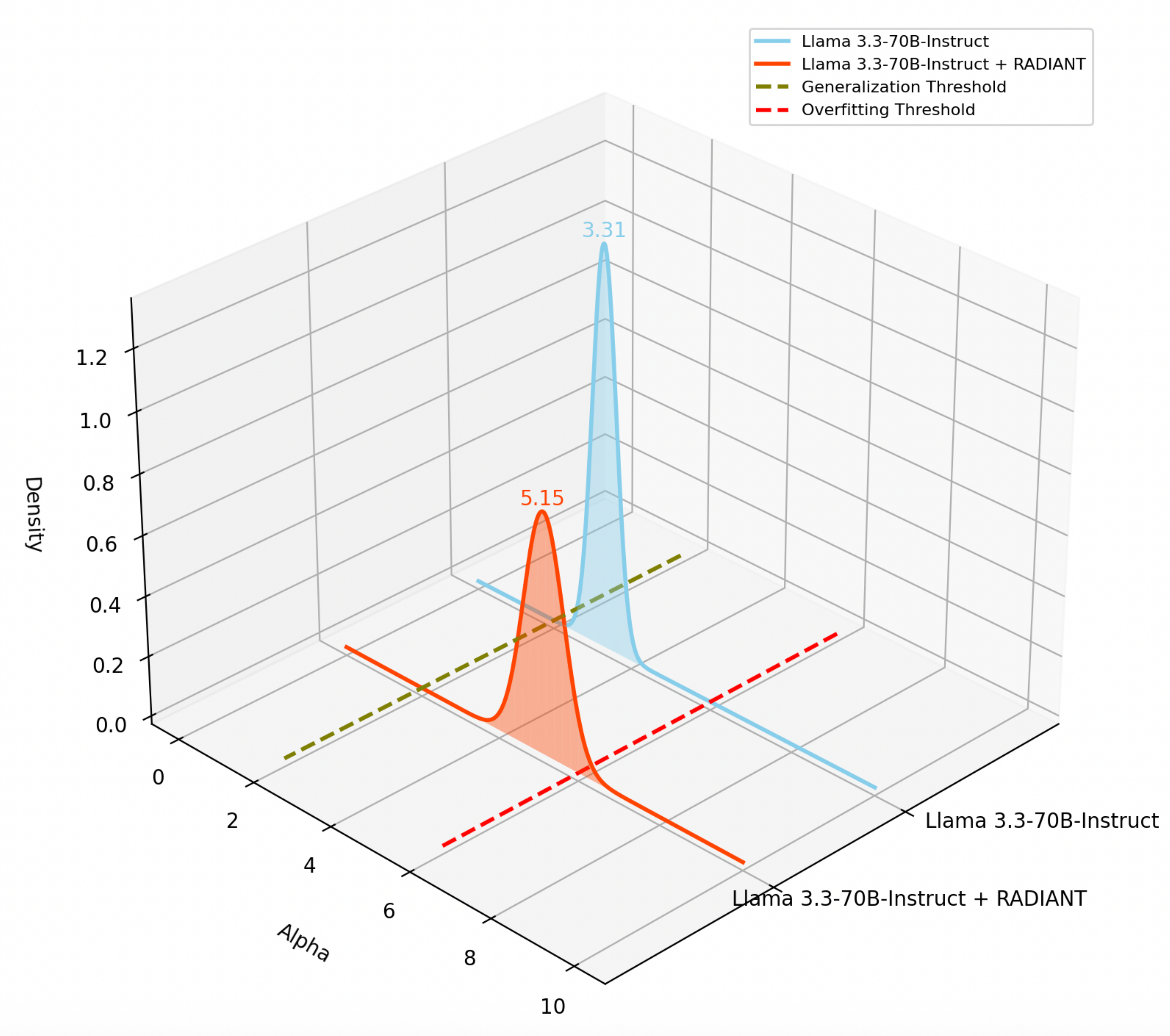}
    \caption{Generalization vs. overfitting trade-off for various DPO-kernels, grounded in Heavy-Tailed Self-Regularization (HTSR) theory. Smaller \(\alpha\) values indicate stronger self-regularization and better generalization, while larger \(\alpha\) values signal overfitting or under-optimized layers.} 
    \label{fig:htsr_generalization_main}
    \vspace{-4mm}
\end{figure}

The \textit{Weighted Alpha} metric \cite{martin2021predicting} offers a novel way to assess generalization and overfitting in LLMs without requiring training or test data. Rooted in Heavy-Tailed Self-Regularization (HT-SR) theory, it analyzes the eigenvalue distribution of weight matrices, modeling the Empirical Spectral Density (ESD) as a power-law \(\rho(\lambda) \propto \lambda^{-\alpha}\). Smaller \(\alpha\) values indicate stronger self-regularization and better generalization, while larger \(\alpha\) values signal overfitting. The \textbf{Weighted Alpha} \(\hat{\alpha}\) is computed as:
$\hat{\alpha} = \frac{1}{L} \sum_{l=1}^L \alpha_l \log \lambda_{\max,l}$,
where \(\alpha_l\) and \(\lambda_{\max,l}\) are the power-law exponent and largest eigenvalue of the \(l\)-th layer, respectively. This formulation highlights layers with larger eigenvalues, providing a practical metric to diagnose generalization and overfitting tendencies. Results reported in \cref{fig:htsr_generalization_main}.


\begin{tcolorbox}[
left=2pt,right=2pt,colback=Navy!5!white,colframe=Navy!75!black,colbacktitle=Navy,
  title=\footnotesize \fontfamily{qbk} \selectfont \textbf{Do aligned LLMs lose generalizability and become overfitted?} ]
  
\vspace{-2mm}
Alignment procedures slightly increase overfitting, with a generalization error drift \(|\Delta \mathcal{E}_{\text{gen}}| \leq 0.1\) (within \(\pm 10\%\)), which is 
acceptable.
\end{tcolorbox}

%% file: 7_conclusion.tex
\section{Conclusion and Future Work}
This work enhances factual consistency in RAG settings by integrating retrieved information into LLMs. We introduce \ecd \hspace{-2mm}, a metric that reveals low RAG-ability across most models. To address this, we propose \radiant \hspace{-1mm}, an improved DPO framework for better external knowledge integration. This framework highlights the need for alignment techniques and entity-aware evaluation. 


Future work will explore how enhancing RAG-ability impacts downstream tasks like QA and summarization while expanding to diverse data formats (e.g., tabular data). We also plan to adapt \radiant for multimodal settings, such as Vision-Language Models (VLMs), to reduce hallucinations and improve image content reliability.


%% file: 8_limitations.tex
\newpage
\section{Limitations} \label{sec:limitaions}
While \radiant aims to improve factual consistency in LLM outputs, it introduces potential safety concerns. The framework has not been systematically tested alongside established safety alignment techniques, leaving uncertain compatibility with existing safeguards. \radiant may inadvertently amplify harmful, biased, or adversarial information embedded in retrieved context by prioritizing factual consistency. This raises the risk of generating unsafe or misleading outputs, particularly if the retrieval process is manipulated to introduce malicious content.


\section{Ethical Considerations}

Future work must rigorously evaluate \radiant against adversarial inputs to mitigate these risks and assess its robustness in high-stakes applications. Furthermore, exploring how \radiant can be integrated with safety alignment methods to prevent misuse and uphold ethical standards in AI-assisted content generation is crucial. Ensuring that factual accuracy improvements do not compromise user safety remains a central goal in the responsible development of this framework.

%% file: 10_appendix.tex
\newpage
\appendix
\section{Appendix}
\label{sec:appendix}

\subsection{Ablation Study: Effect of Temperature on Entity-Context Divergence Distribution}

In this section, we present an ablation study to investigate the effect of changing temperature on the Entity-Context Divergence Distribution for various models. The analysis reveals that for a particular model, altering the temperature does not result in significant deviations in the distribution, as the distributions mostly overlap. This finding underscores the stability of divergence metrics across different temperature settings. The corresponding figures for each model are provided below.


\begin{figure}[h!]
    \centering
    \includegraphics[width=\columnwidth]{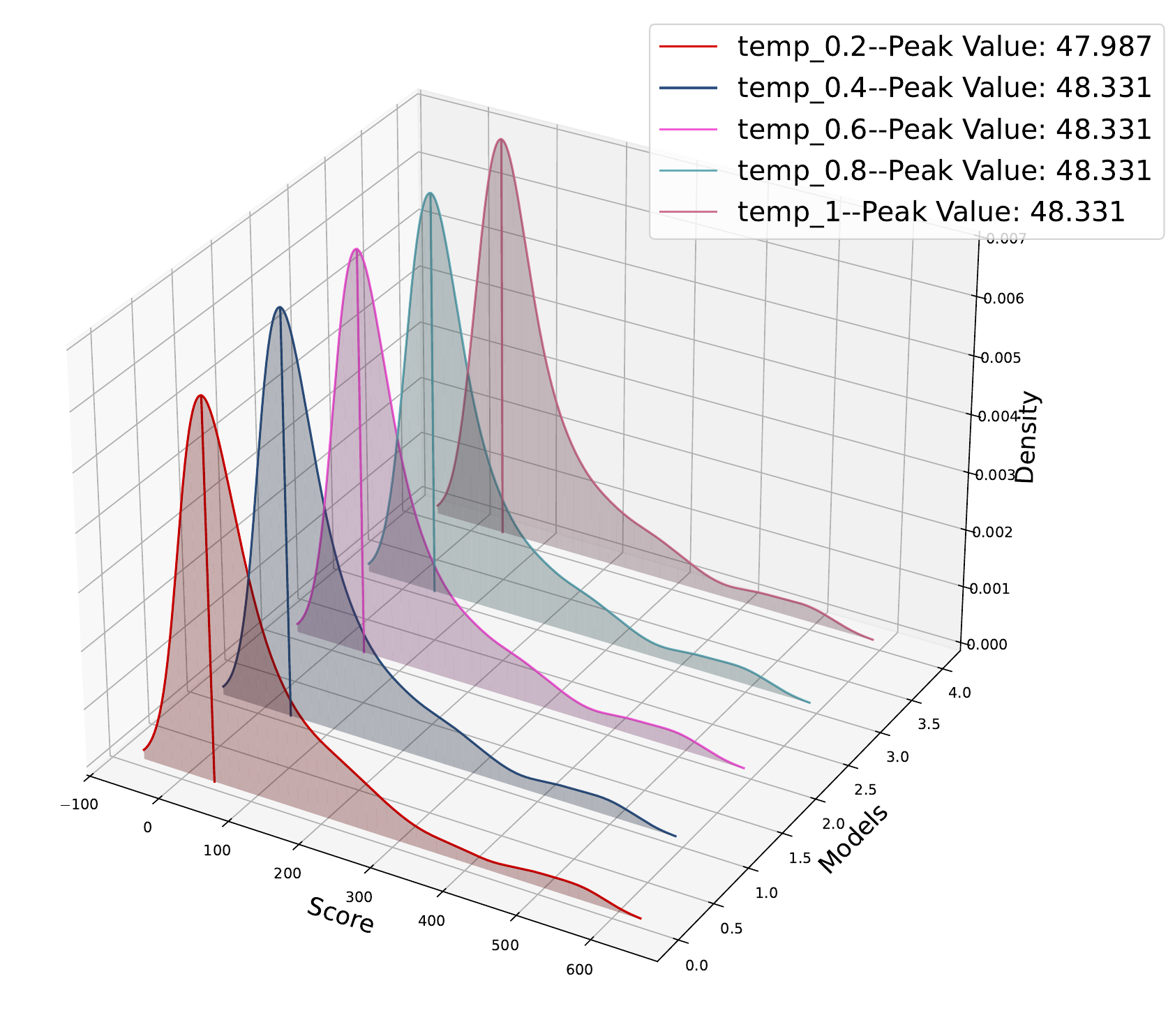}
    \caption{Entity-Context Divergence Distribution for Claude 3.5 Sonnet}
    \label{fig:divergence_distribution in Claude 3.5 sonnet}
\end{figure}

\begin{figure}[h!]
    \centering
    \includegraphics[width=\columnwidth]{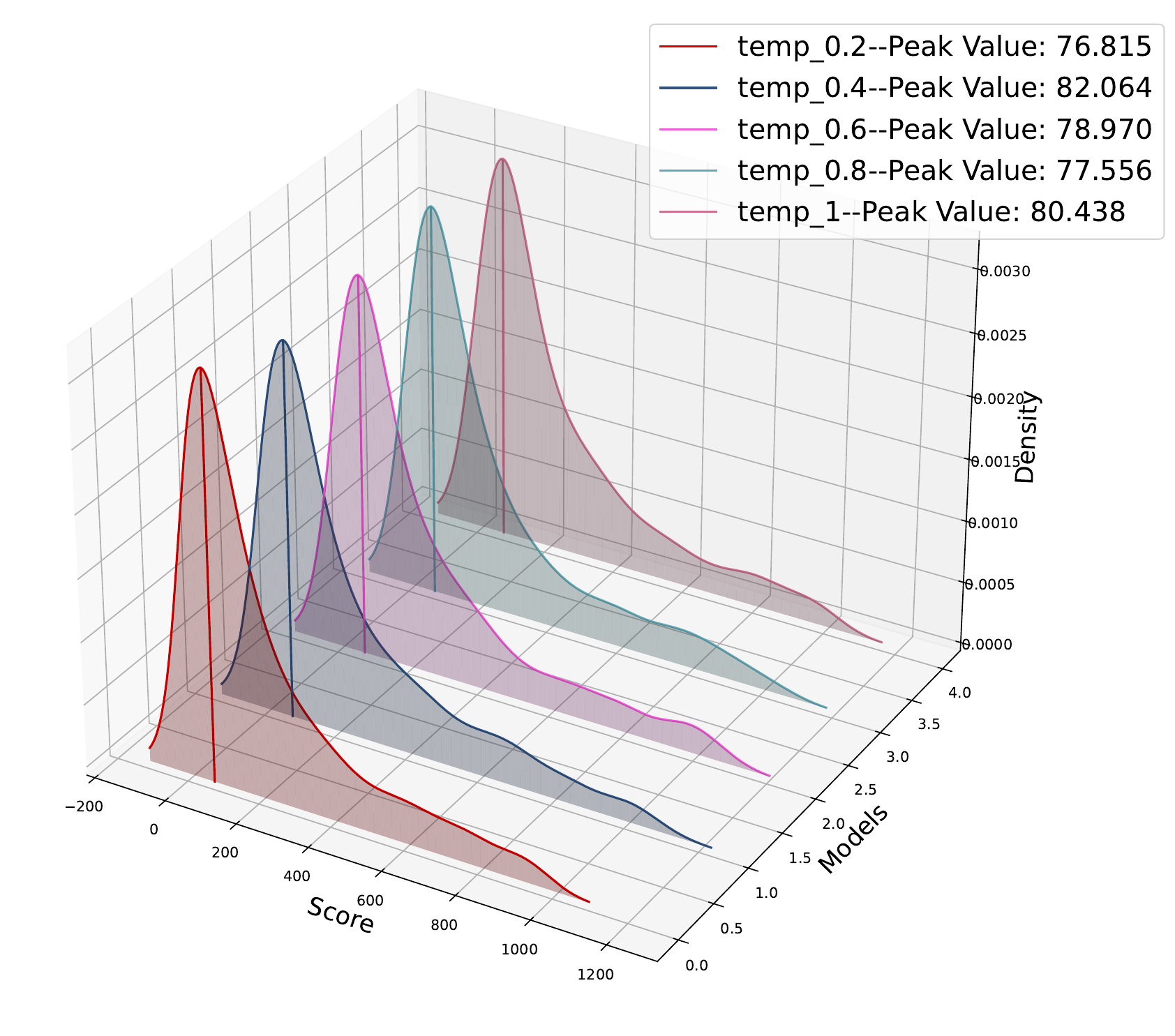}
    \caption{Entity-Context Divergence Distribution for Gemma2 9b}
    \label{fig:divergence_distribution in Gemma2_9b}
\end{figure}

\begin{figure}[h!]
    \centering
    \includegraphics[width=\columnwidth]{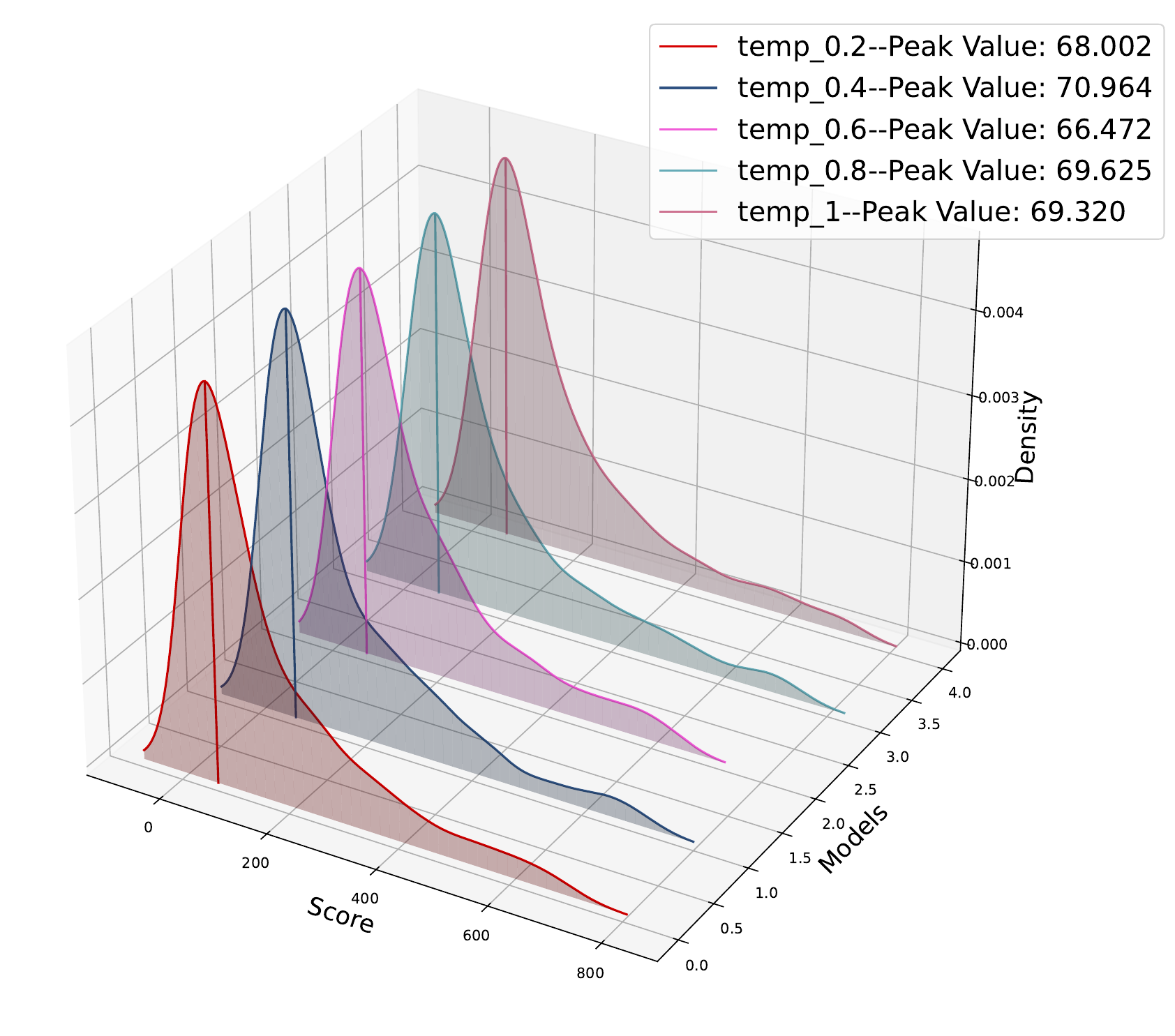}
    \caption{Entity-Context Divergence Distribution for Gemma-2-27b-it}
    \label{fig:divergence_distribution in gemma-2-27b-it}
\end{figure}

\begin{figure}[h!]
    \centering
    \includegraphics[width=\columnwidth]{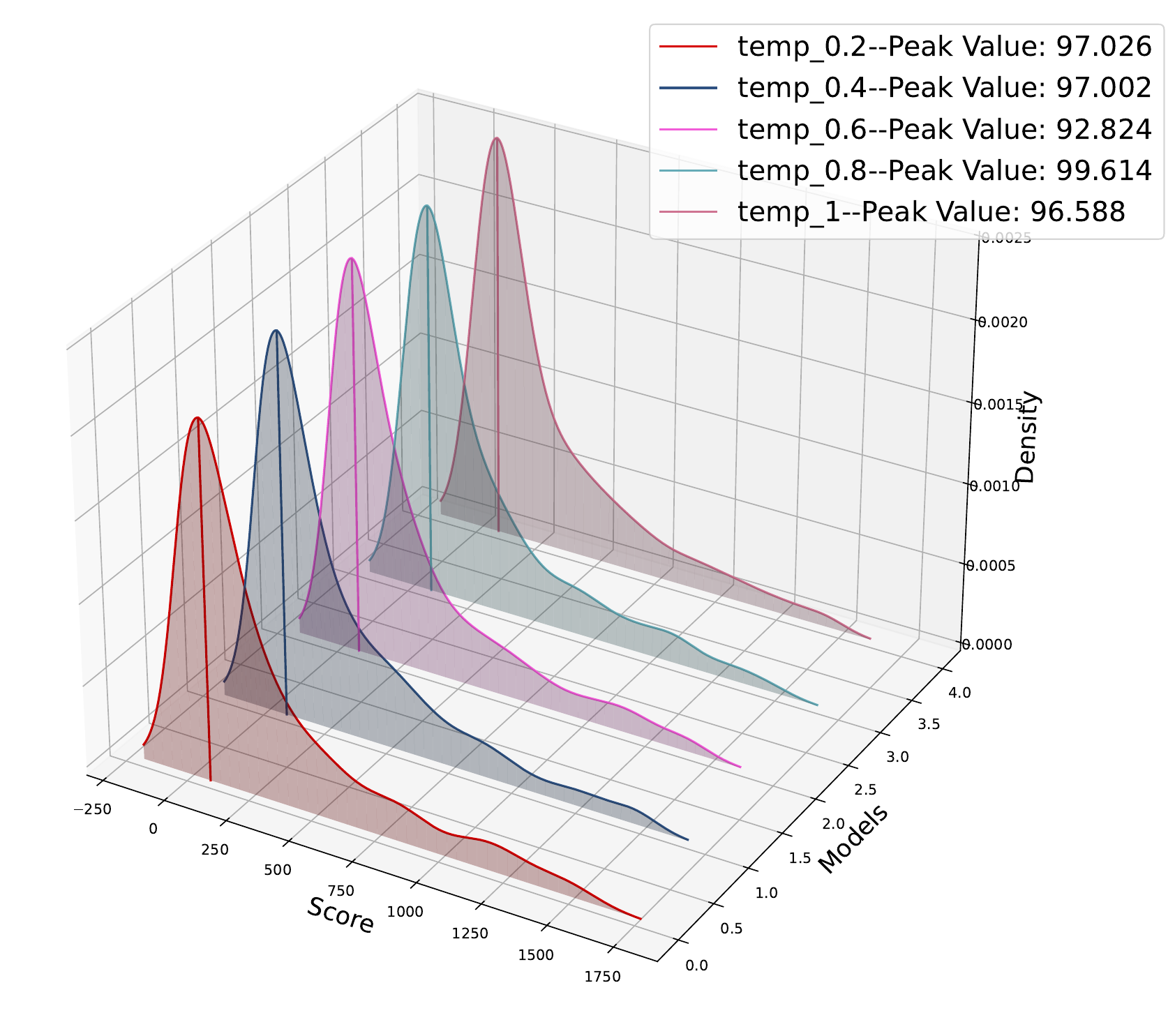}
    \caption{Entity-Context Divergence Distribution for Llama-3.1-70b}
    \label{fig:divergence_distribution in Llama-3.1-70B}
\end{figure}

\begin{figure}[h!]
    \centering
    \includegraphics[width=\columnwidth]{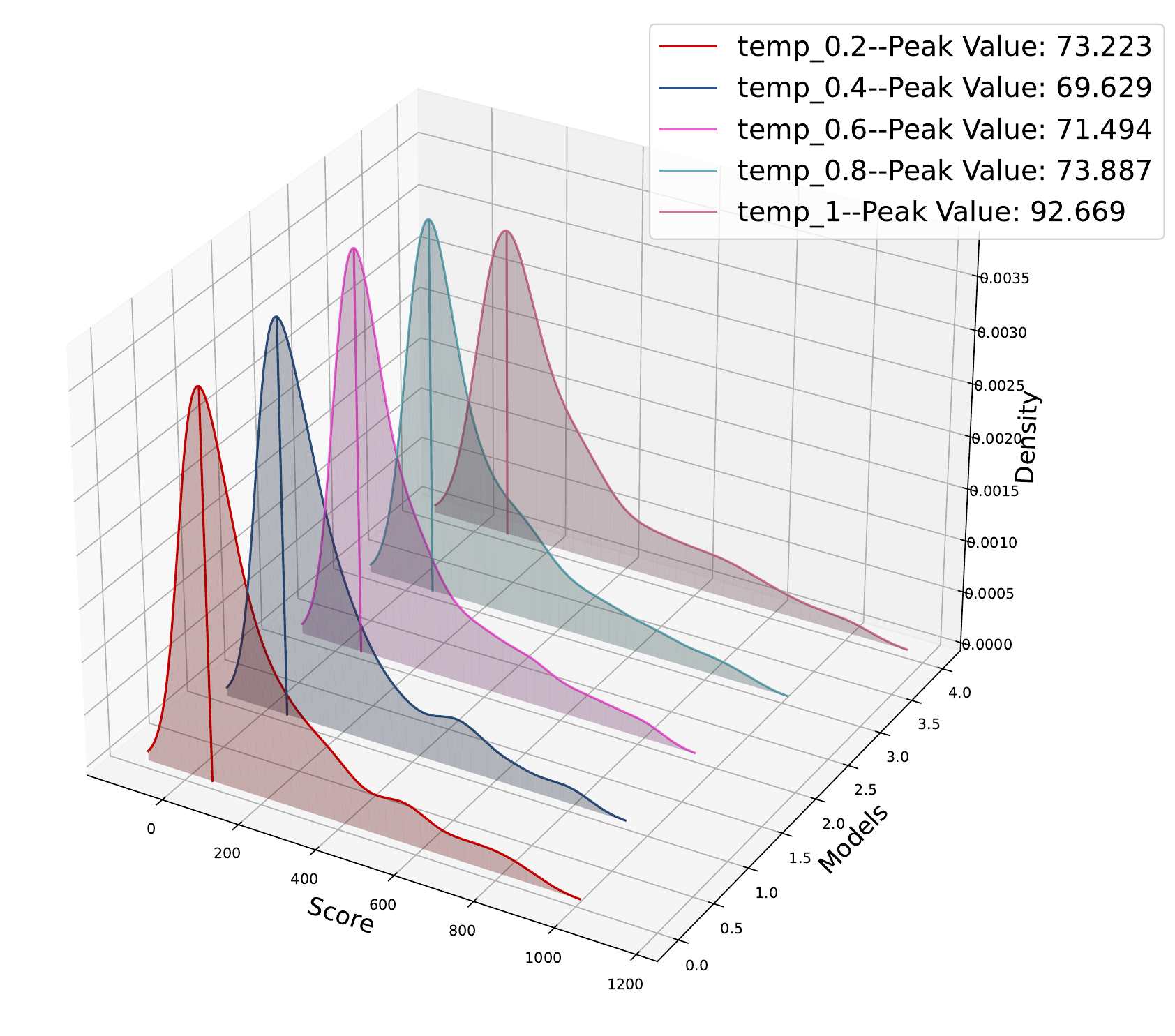}
    \caption{Entity-Context Divergence Distribution for Llama-3.2-3B-Instruct-Turbo}
    \label{fig:divergence_distribution in Llama-3.2-3B-Instruct-Turbo}
\end{figure}

\begin{figure}[h!]
    \centering
    \includegraphics[width=\columnwidth]{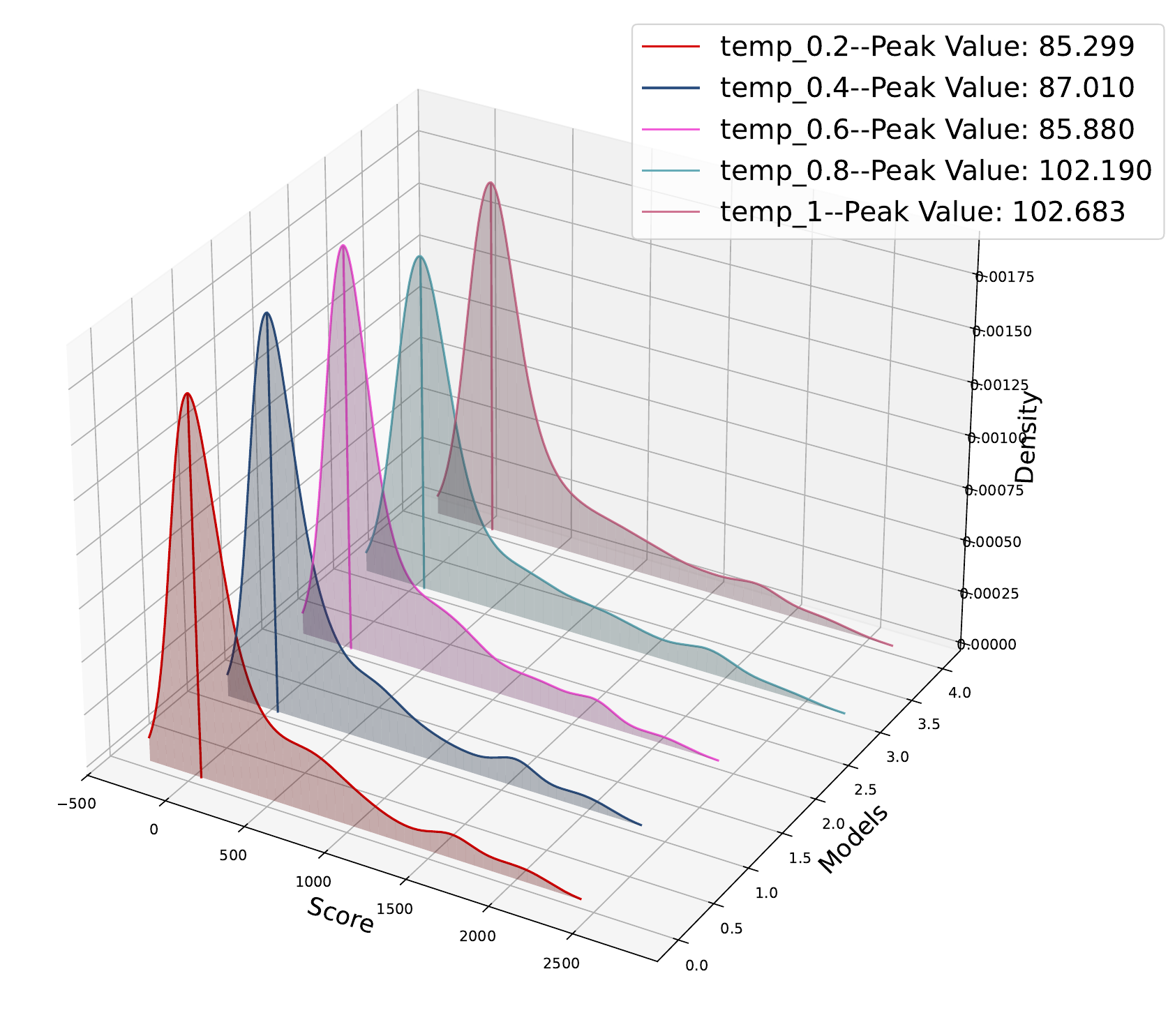}
    \caption{Entity-Context Divergence Distribution for Mixtral-8x7b-instruct}
    \label{fig:divergence_distribution in Mixtral-8x7b-instruct}
\end{figure}

\begin{figure}[h!]
    \centering
    \includegraphics[width=\columnwidth]{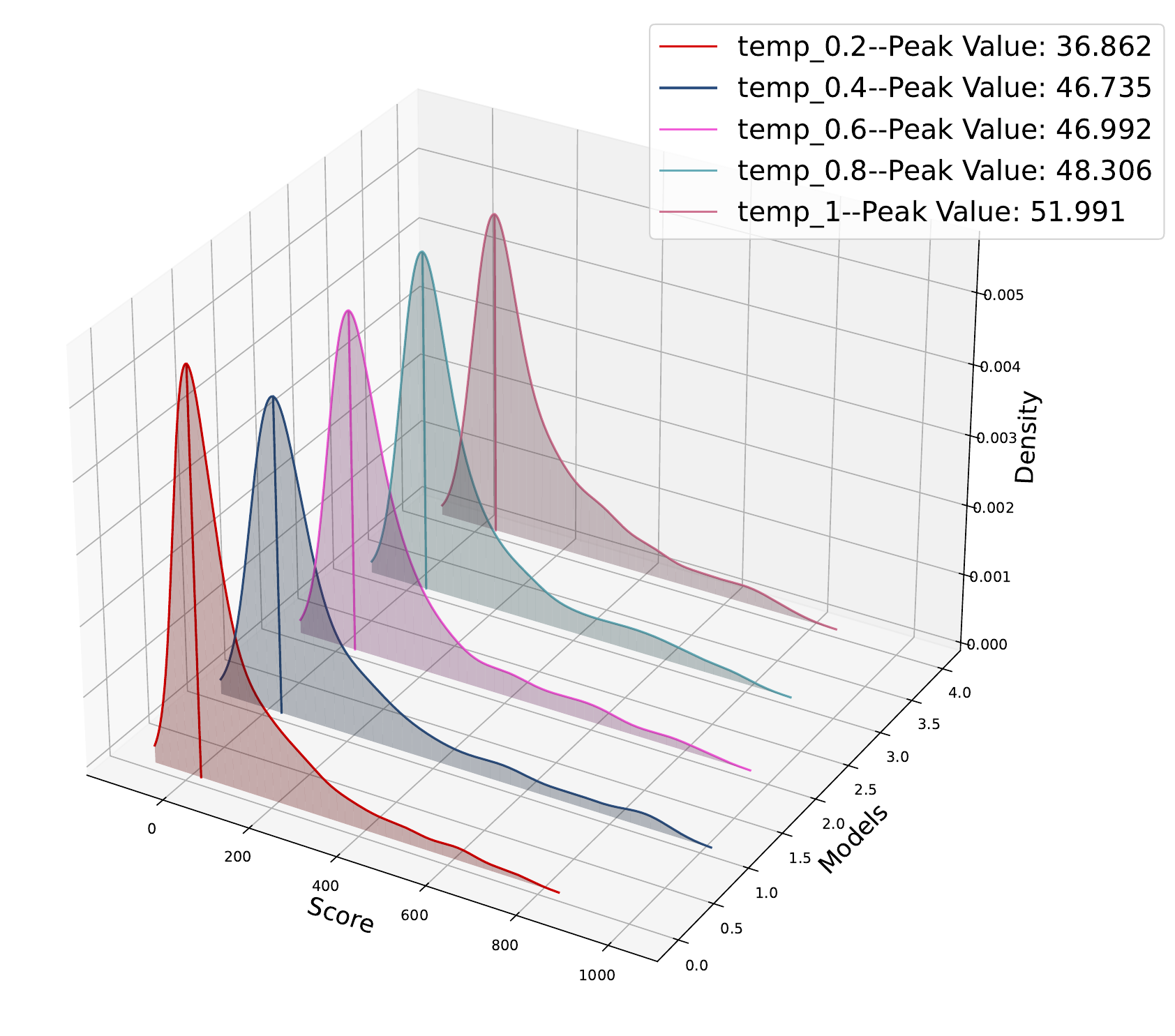}
    \caption{Entity-Context Divergence Distribution for Qwen2-72b-instruct}
    \label{fig:divergence_distribution in Qwen2-72b-instruct}
\end{figure}


\begin{figure}[h!]
    \centering
    \includegraphics[width=\columnwidth]{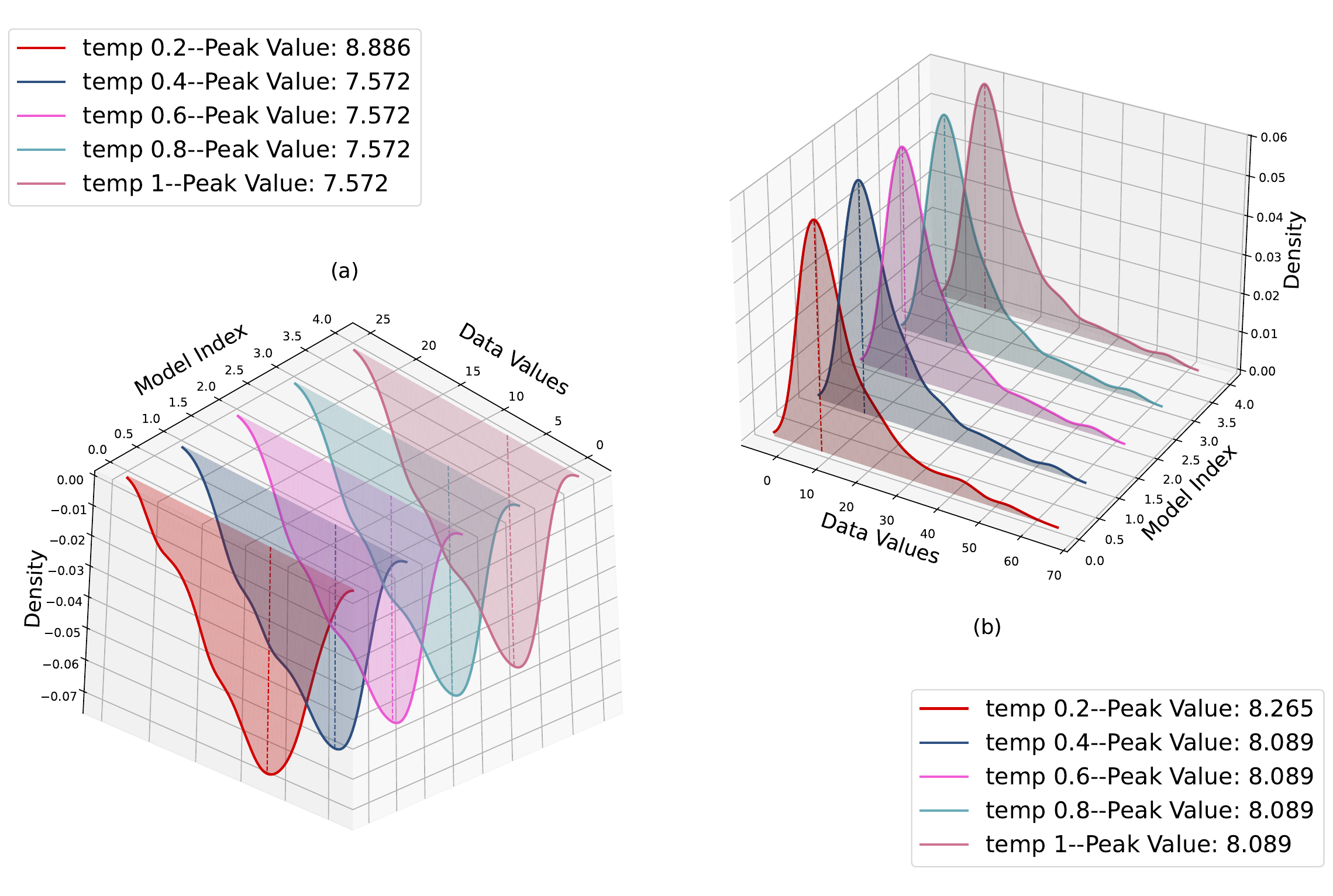}
    \caption{Entity-Context Divergence Distribution for Claude 3.5 Sonnet
    (a) Extra Entities
    (b) Missing Entities.}
    
    \label{fig:divergence_distribution_temp in Claude 3.5 sonnet}
\end{figure}

\begin{figure}[h!]
    \centering
    \includegraphics[width=\columnwidth]{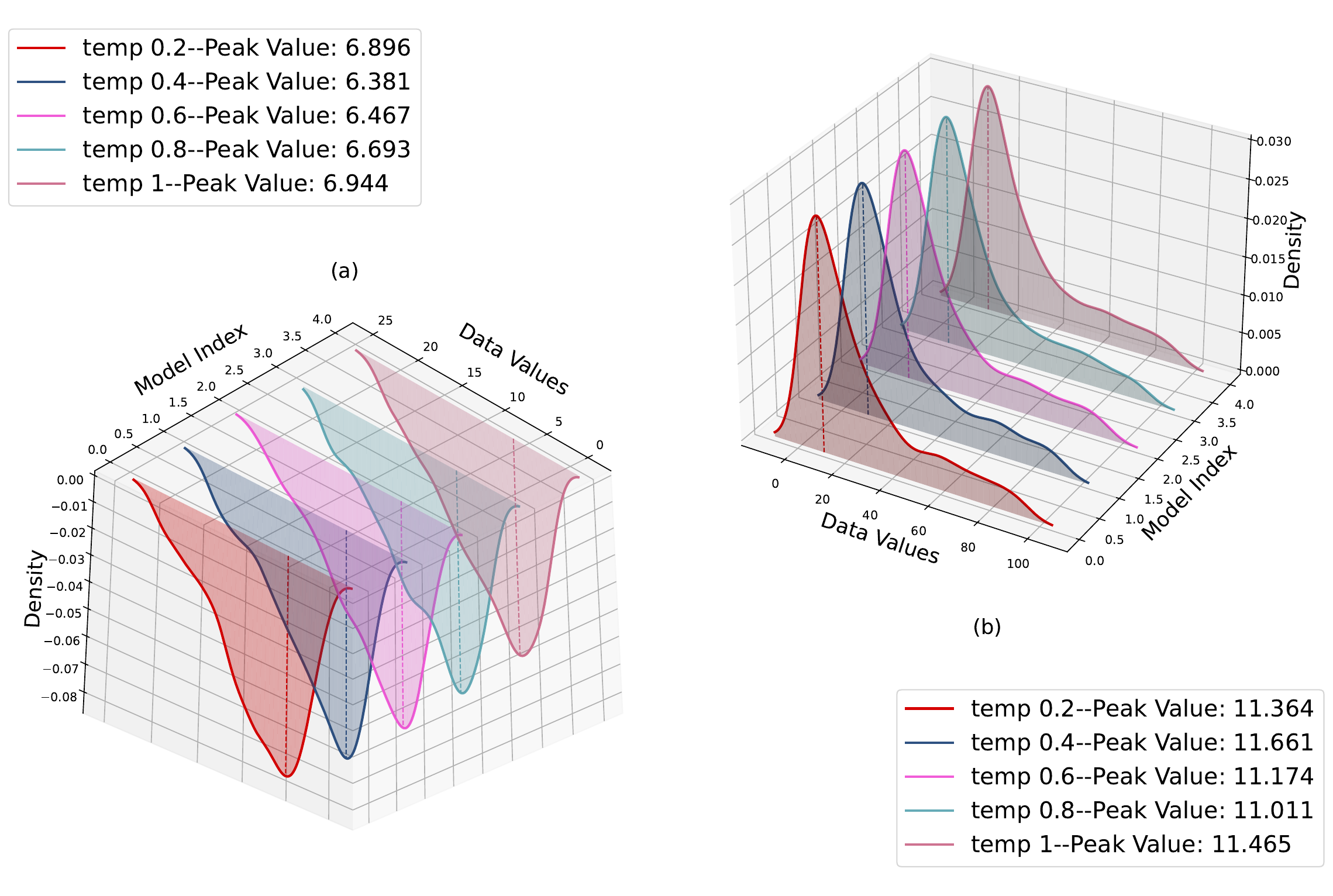}
    \caption{Entity-Context Divergence Distribution for Gemma2 9b
    (a) Extra Entities
    (b) Missing Entities.}
    \label{fig:divergence_distribution_temp in Gemma2_9b}
\end{figure}

\begin{figure}[h!]
    \centering
    \includegraphics[width=\columnwidth]{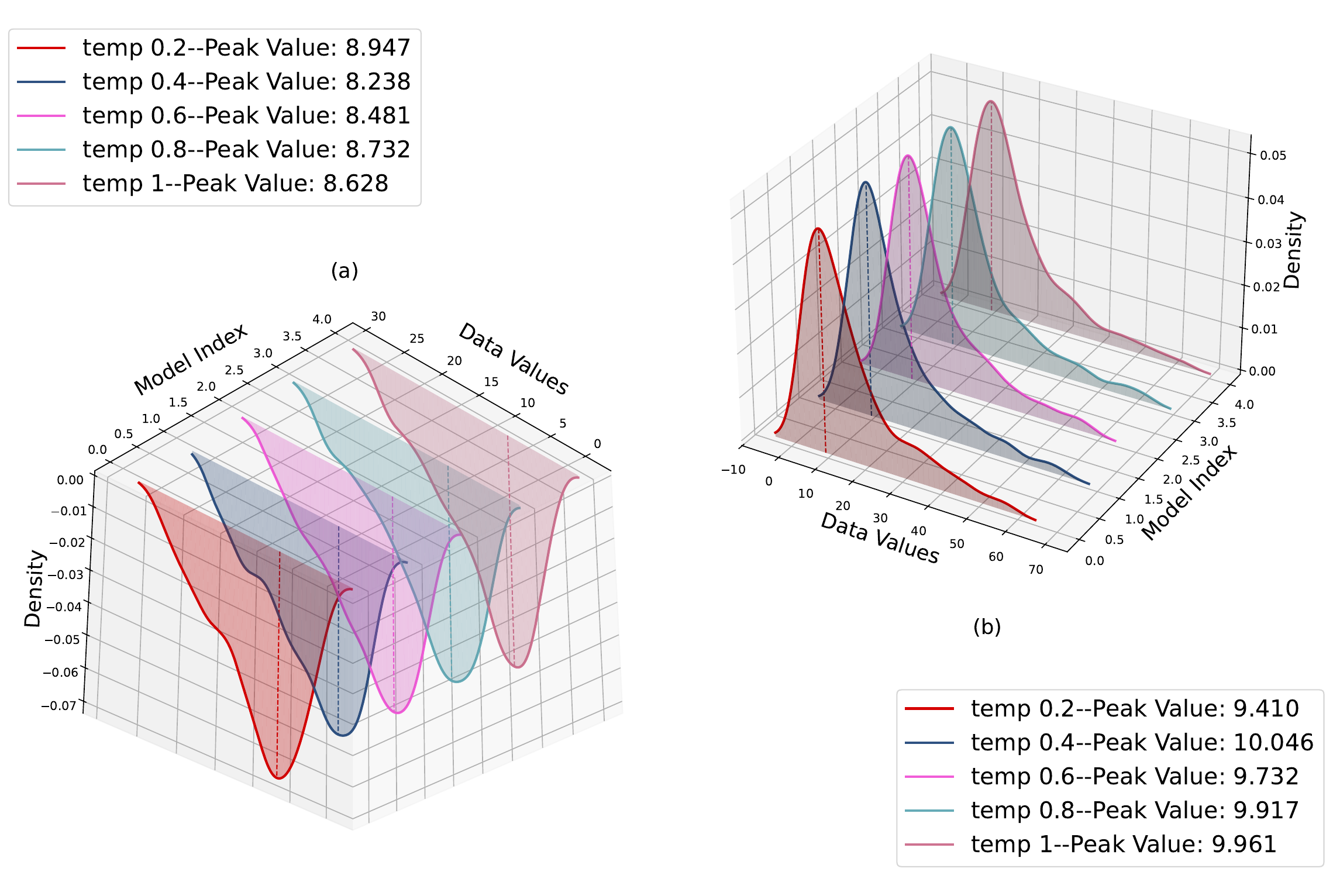}
    \caption{Entity-Context Divergence Distribution for Gemma-2-27b-it
    (a) Extra Entities
    (b) Missing Entities.}
    \label{fig:divergence_distribution_temp in gemma-2-27b-it}
\end{figure}

\begin{figure}[h!]
    \centering
    \includegraphics[width=\columnwidth]{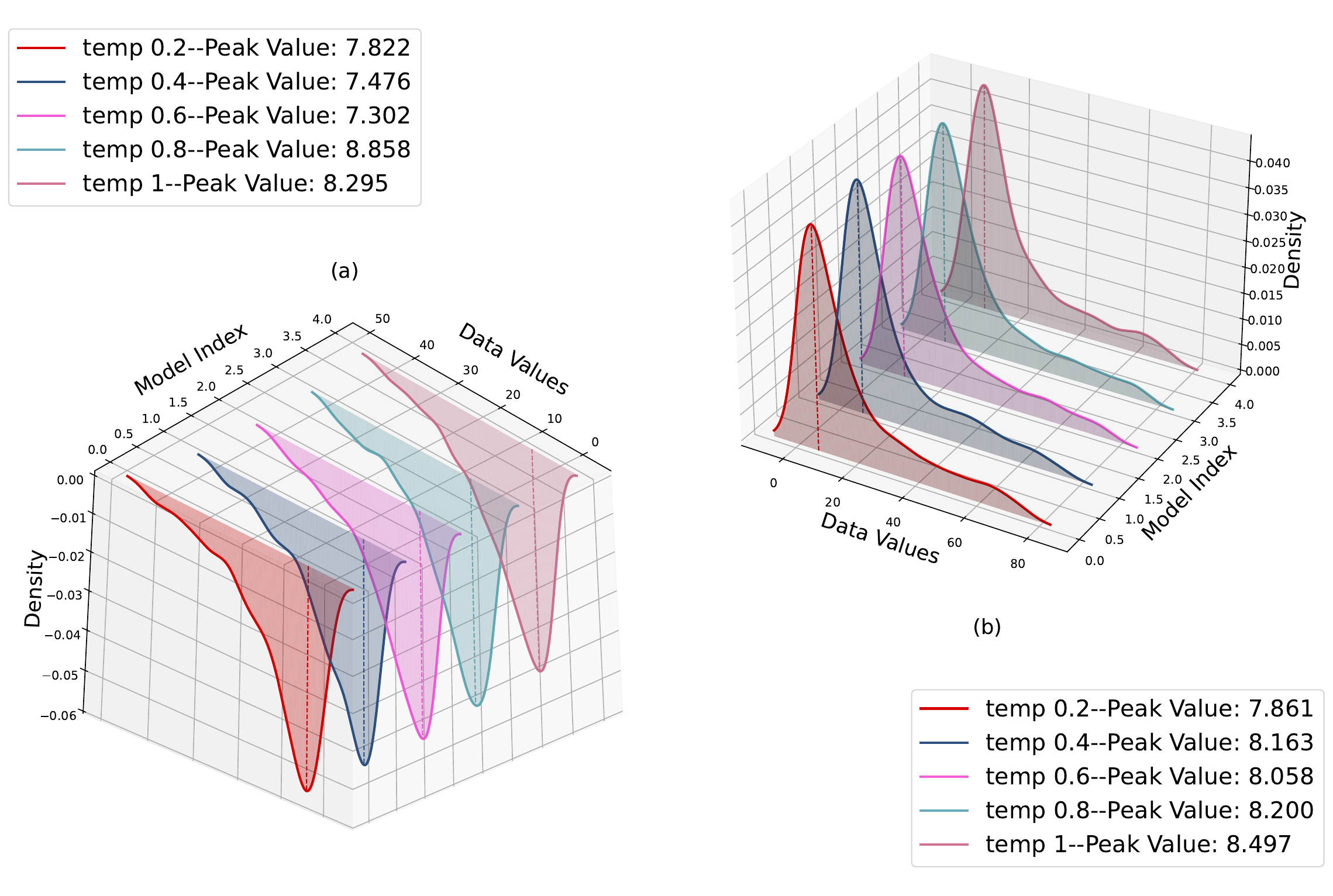}
    \caption{Entity-Context Divergence Distribution for Llama-3.1-70b
    (a) Extra Entities
    (b) Missing Entities.}
    \label{fig:divergence_distribution_temp in Llama-3.1-70B}
\end{figure}

\begin{figure}[h!]
    \centering
    \includegraphics[width=\columnwidth]{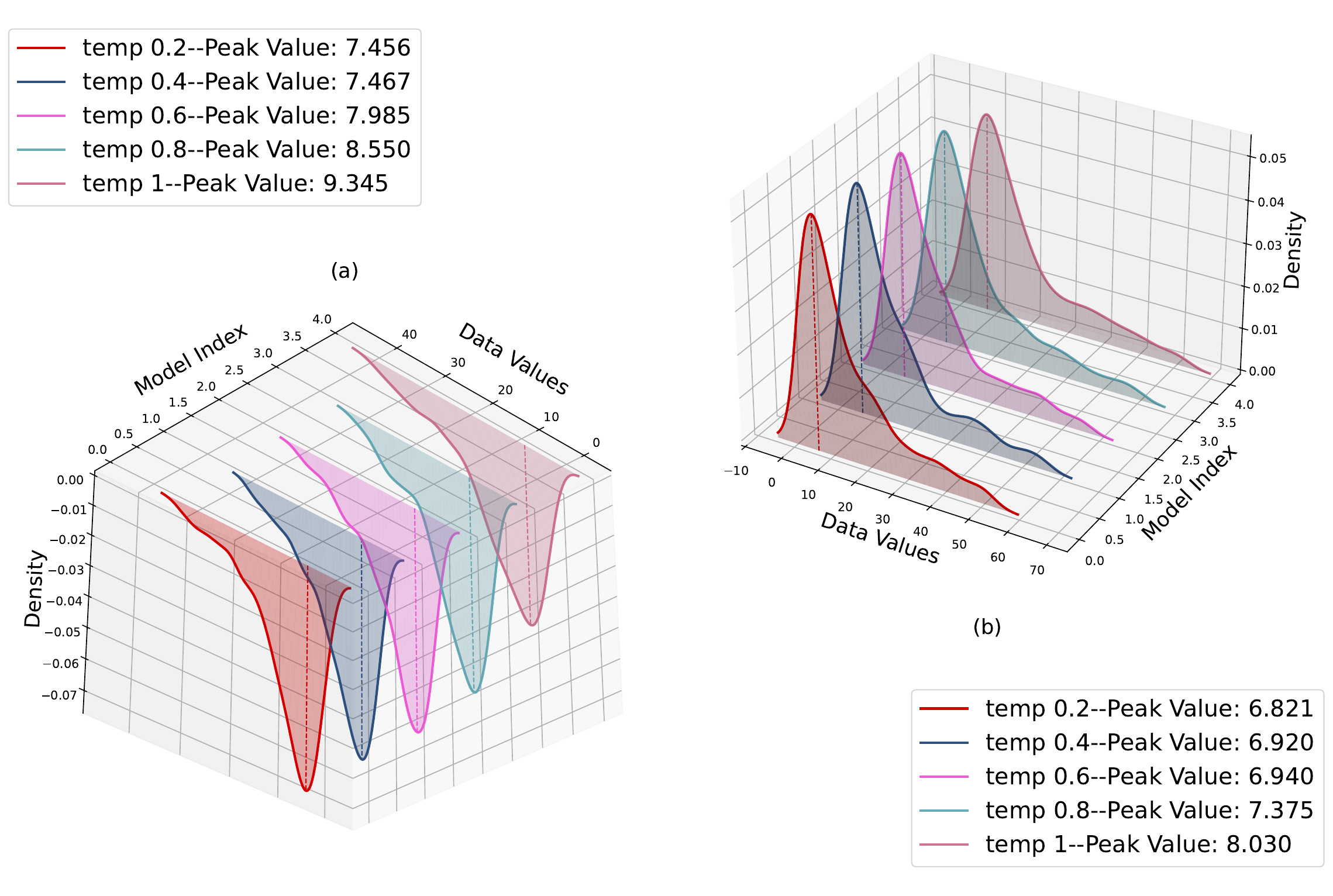}
    \caption{Entity-Context Divergence Distribution for Llama-3.2-3B-Instruct-Turbo
    (a) Extra Entities
    (b) Missing Entities.}
    \label{fig:divergence_distribution_temp in Llama-3.2-3B-Instruct-Turbo}
\end{figure}

\begin{figure}[h!]
    \centering
    \includegraphics[width=\columnwidth]{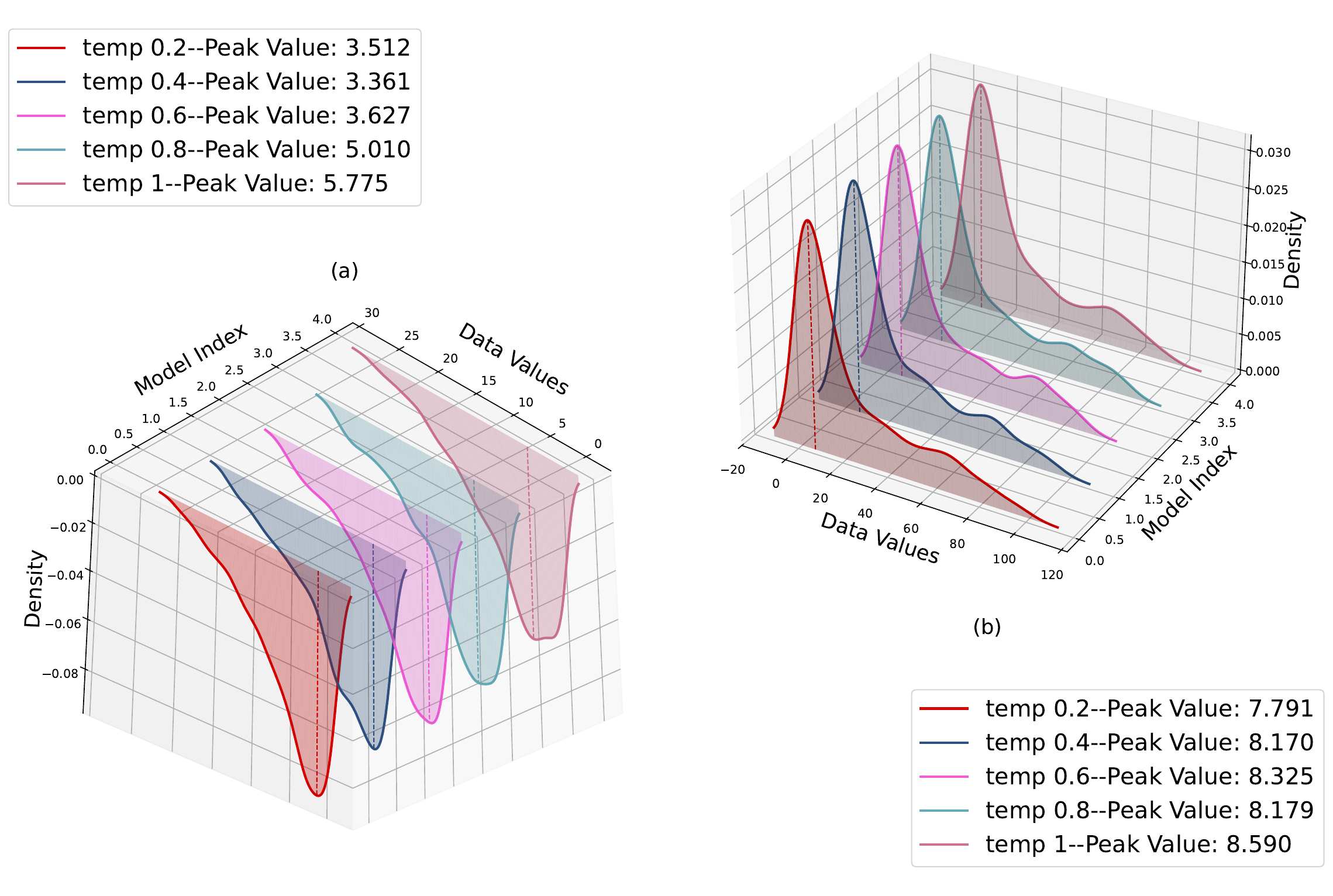}
    \caption{Entity-Context Divergence Distribution for Mixtral-8x7b-instruct
    (a) Extra Entities
    (b) Missing Entities.}
    \label{fig:divergence_distribution_temp in Mixtral-8x7b-instruct}
\end{figure}

\begin{figure}[h!]
    \centering
    \includegraphics[width=\columnwidth]{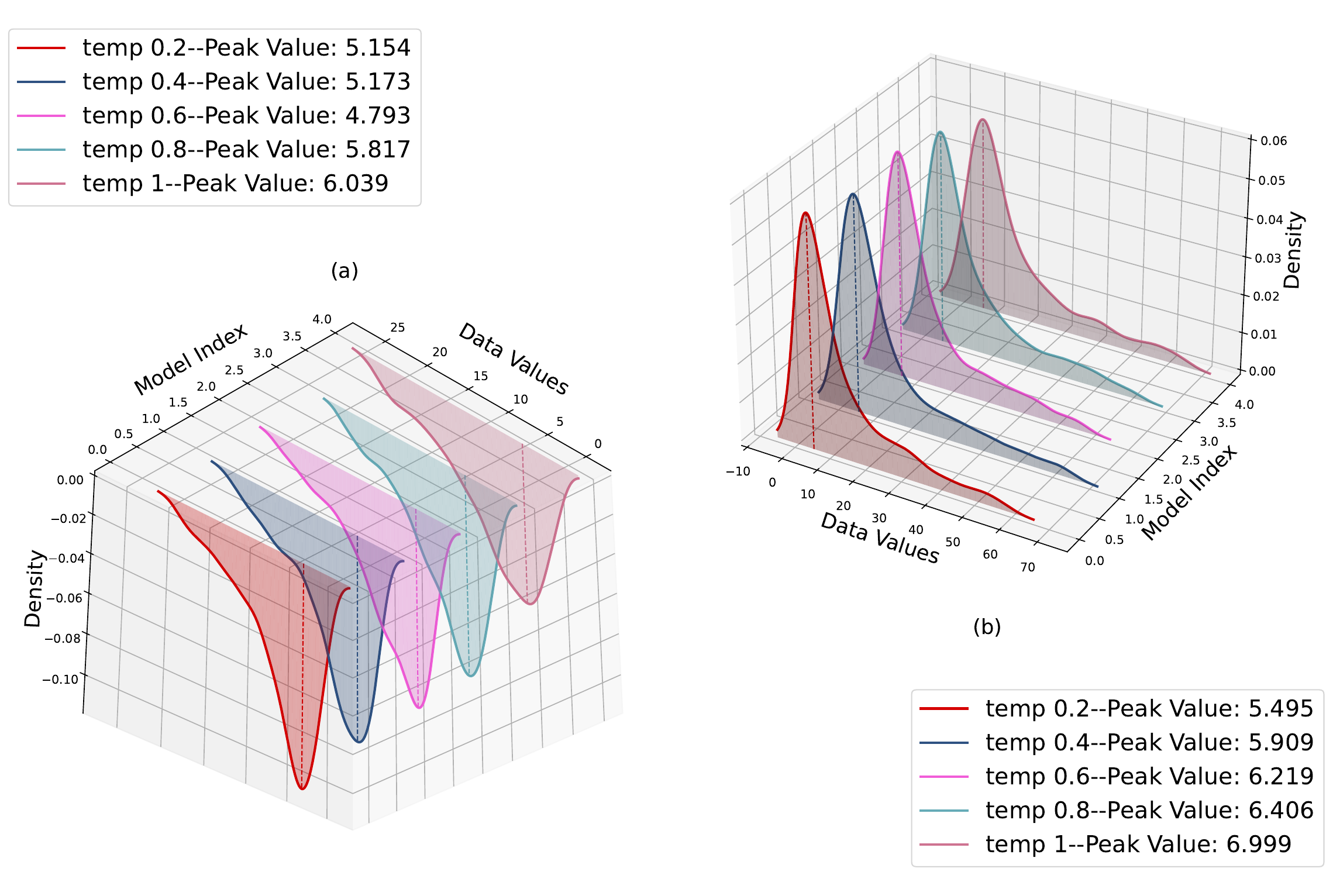}
    \caption{Entity-Context Divergence Distribution for Qwen2-72b-instruct
    (a) Extra Entities
    (b) Missing Entities.}
    \label{fig:divergence_distribution_temp in Qwen2-72b-instruct}
\end{figure}

\section{KDE Plots for Extra and Missing Entities Across Scenarios}

The following figures illustrate KDE plots for extra and missing entities across four distinct scenarios: 
(i) Vanilla Generation Without Context, (ii) Generation With Perfect Context, 
(iii) Generation With Web-Retrieved Context, and (iv) Generation With Synthesized Context. 

These plots represent results obtained from seven models: Gemini-1.5-pro, Gemini-1.0-pro, 
Gemma-2-9b, LLAMA 3 8b, 
and Qwen2-72B. Each figure corresponds to a specific scenario and visualizes the differences 
in performance based on the KDE distribution of scores for missing and extra entities.


\begin{figure}[h!]
    \centering
    \includegraphics[width=\columnwidth]{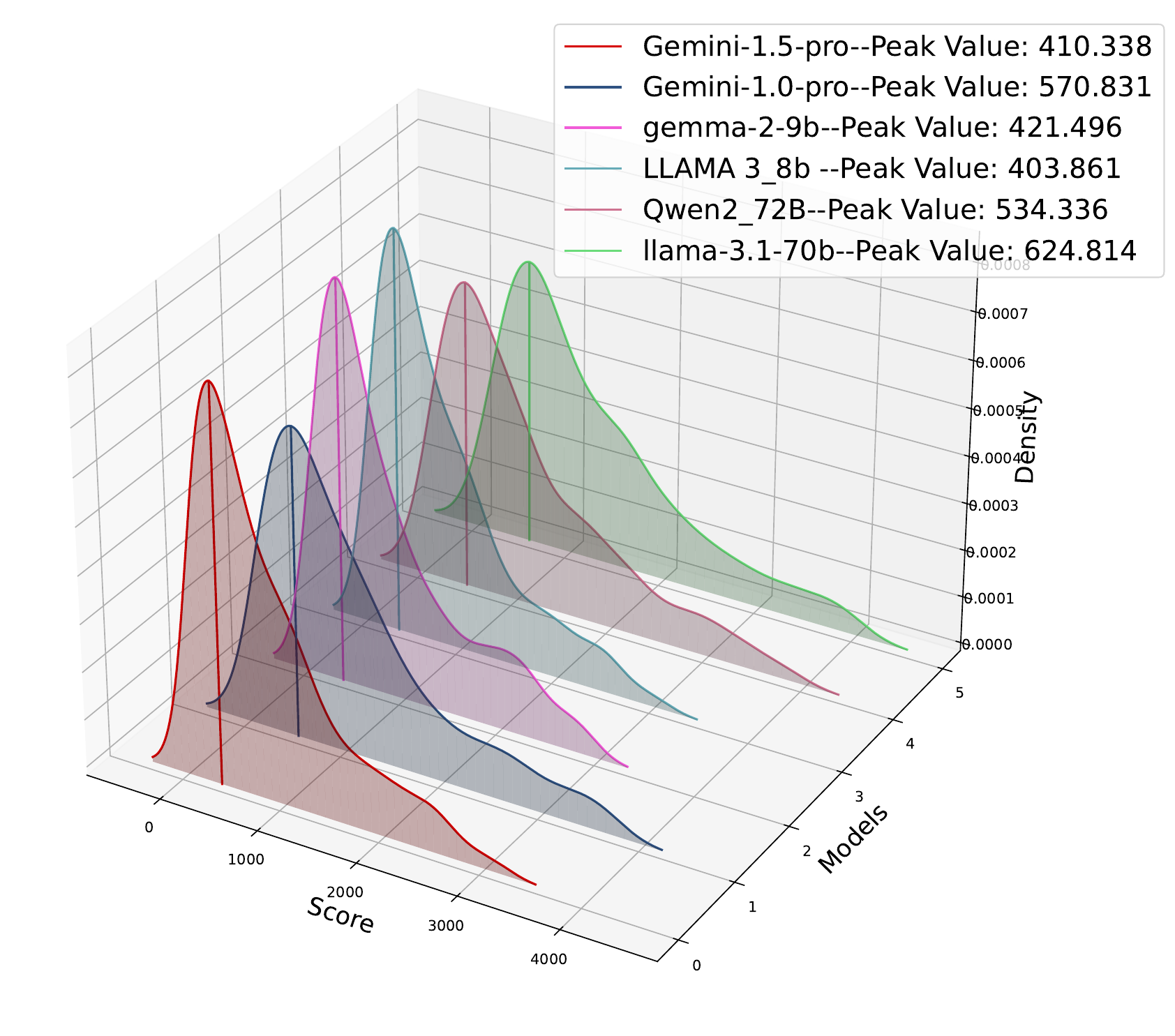}
    \caption{Entity-Context Divergence Distribution for the AI Text
    Generation Without any Context 
}
    \label{fig:vanilla_generation_scenario}
\end{figure}

\begin{figure}[h!]
    \centering
    \includegraphics[width=\columnwidth]{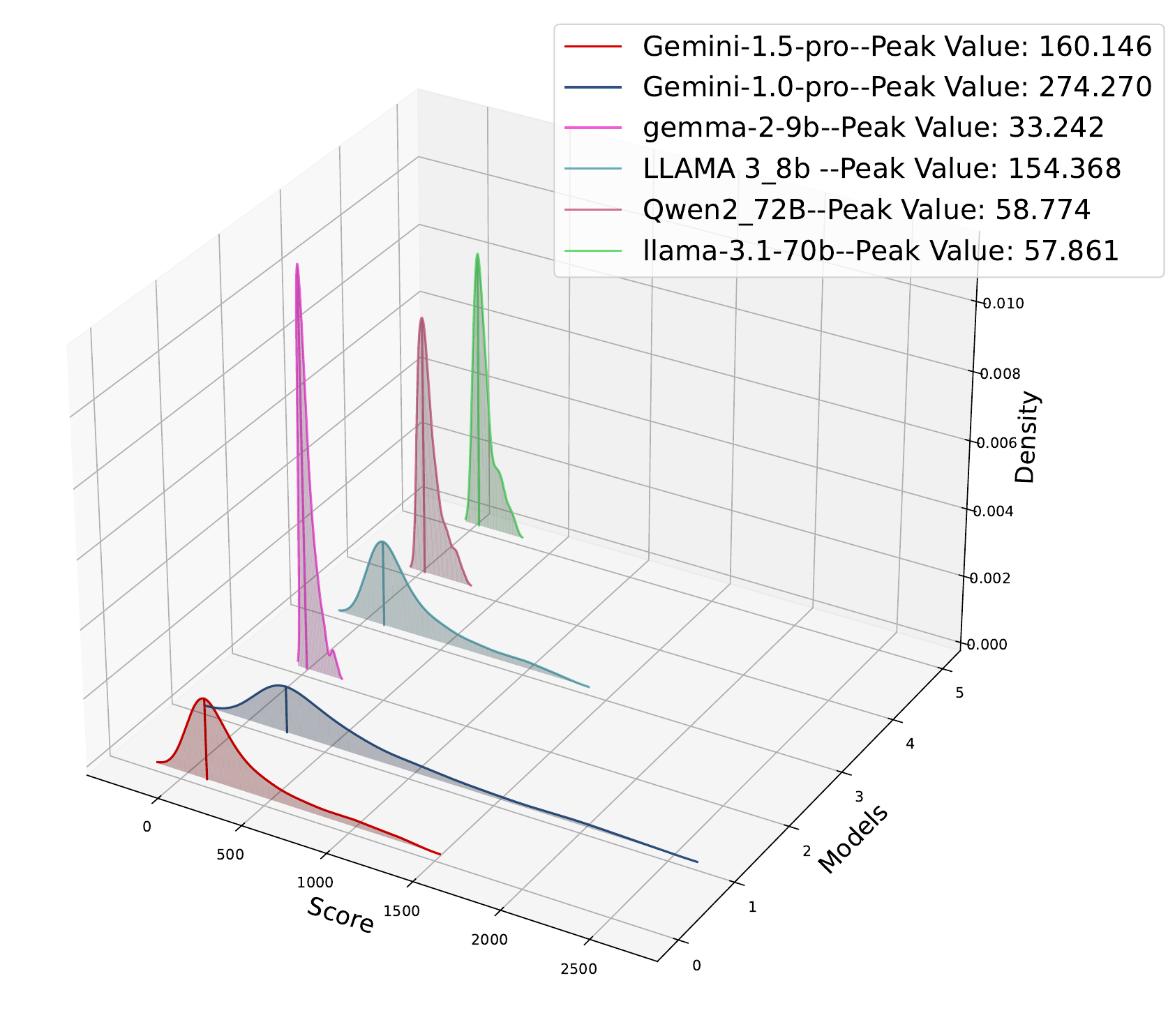}
    \caption{Entity-Context Divergence Distribution for the AI Text
    Generation With Perfect Context, i.e. Twitter articles
    }
    \label{fig:vanilla_generation_scenario}
\end{figure}

\begin{figure}[h!]
    \centering
    \includegraphics[width=\columnwidth]{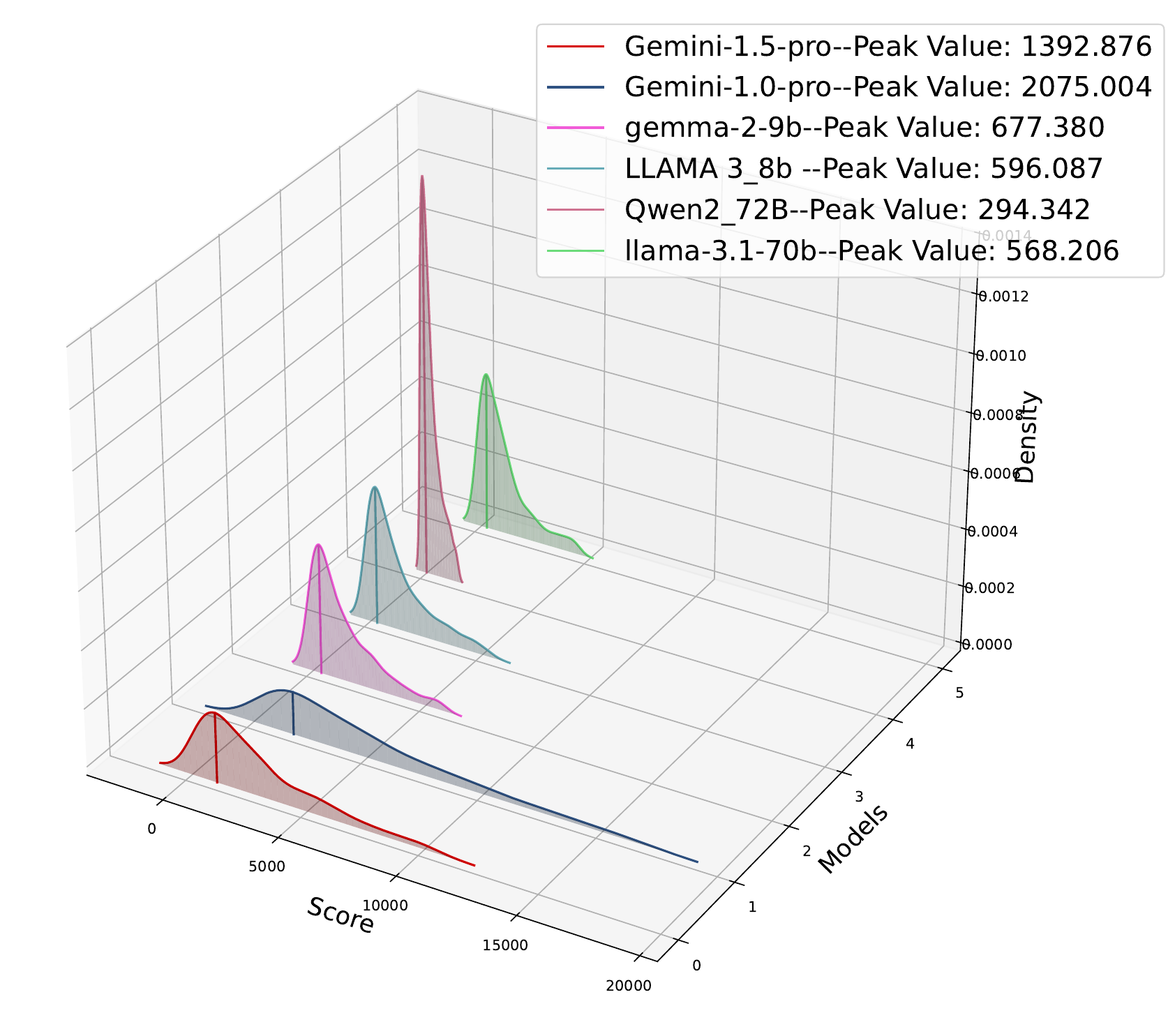}
    
    \caption{Entity-Context Divergence Distribution for the AI Text
    Generation With Web-Retrieved Context 
    }
    
    \label{fig:Web-Retrieved_generation_scenario}
\end{figure}

\begin{figure}[h!]
    \centering
    \includegraphics[width=\columnwidth]{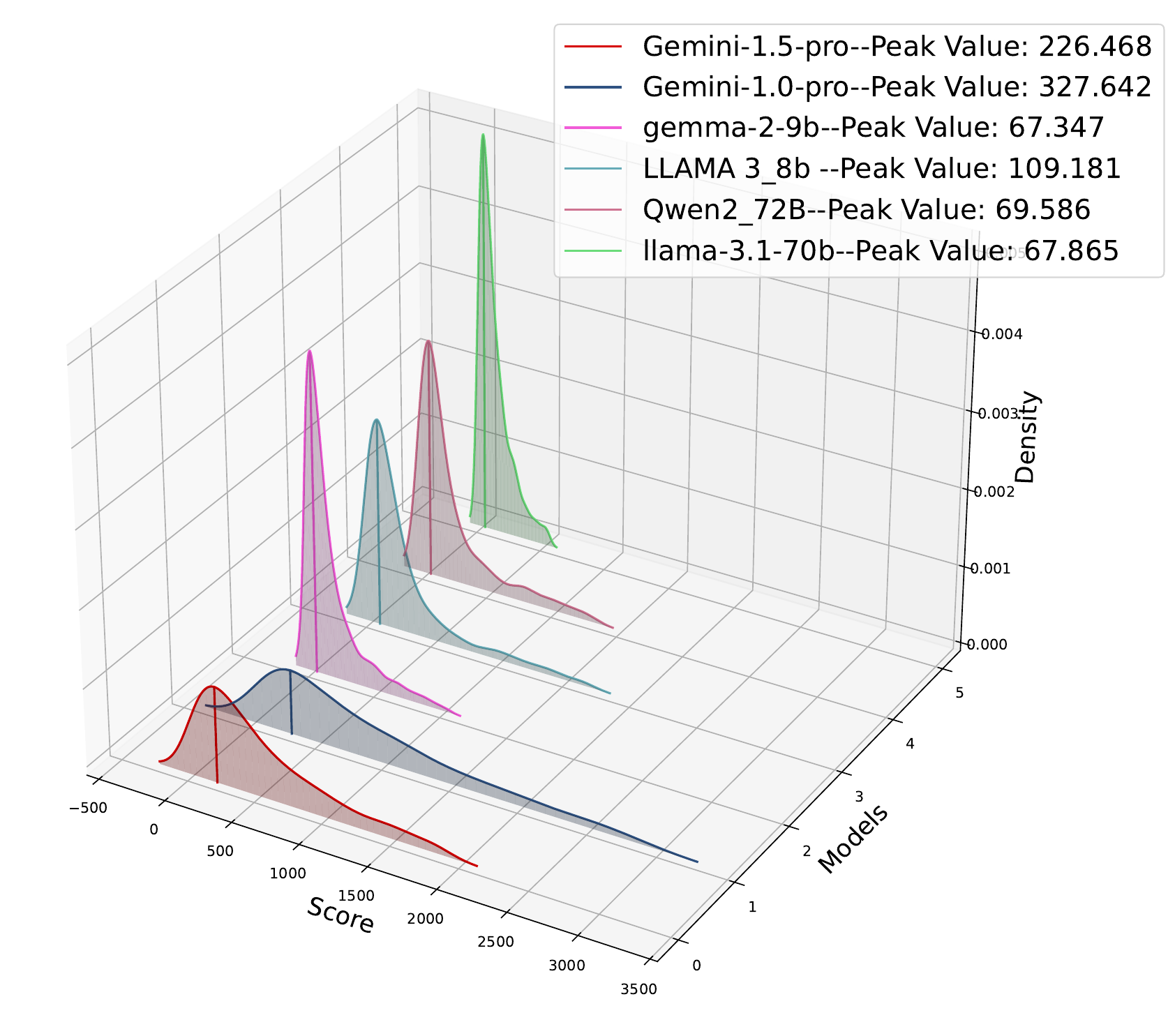}
    \caption{Entity-Context Divergence Distribution for the AI Text
    Generation With Synthesized Context, i.e. injecting entity replacement to make the context factually incorrect
    }
    \label{fig:synthesized_generation_scenario}
\end{figure}

The plots help to visualize how the models perform under each scenario, providing insights into the alignment and divergence of generated content with the expected entity distributions.

\subsection{Example Calculation of ECD}
To illustrate the BoW-based Entity-Context Divergence (ECD) calculation, consider the following example:

\paragraph{Example Setup}
Suppose we have the following retrieved and AI-generated contexts:
\begin{itemize}
    \item \textbf{Retrieved Context} ($C_r$): \{"Modi", "India", "BJP", "election", "leader"\}
    \item \textbf{AI-Generated Context} ($C_g$): \{"Modi", "India", "G7", "summit", "economy"\}
\end{itemize}

\paragraph{Step 1: Identify Shared, Missing, and Added Entities}
\begin{itemize}
    \item \textbf{Shared Entities} ($E_{\text{common}}$): \{"Modi", "India"\}
    \item \textbf{Missing Entities} ($E_{\text{missing}}$): \{"BJP", "election", "leader"\}
    \item \textbf{Added Entities} ($E_{\text{added}}$): \{"G7", "summit", "economy"\}
\end{itemize}

\paragraph{Step 2: Calculate Contextual Divergence for Shared Entities}
For shared entity "Modi", assume $W_r(e) = \{\text{election, BJP, India}\}$ and $W_g(e) = \{\text{G7, summit, economy}\}$. The Jaccard divergence is:
\begin{equation}
    d_{\text{Jaccard}}(W_r(\text{Modi}), W_g(\text{Modi})) = 1 - \frac{0}{6} = 1
\end{equation}

\paragraph{Step 3: Calculate Penalties for Missing and Added Entities}
\begin{itemize}
    \item \textbf{Missing Entities Penalty (ME)}: Assume ranks are \{2, 3, 4\} and $\sigma = 0.5$. Then, \( ME(C_r, C_g) = \frac{(2 + 3 + 4) \cdot 0.5}{2} = 4.5 \).
    \item \textbf{Added Entities Penalty (AE)}: Assume ranks are \{2, 3, 4\} and $\sigma = 0.5$. Then, \( AE(C_r, C_g) = \frac{(2 + 3 + 4) \cdot 0.5}{2} = 4.5 \).
\end{itemize}

\paragraph{Final ECD Calculation}
\begin{equation}
    ECD(C_r, C_g) = 1 + 4.5 + 4.5 = 10
\end{equation}

This example demonstrates how to compute ECD using BoW-based measures for contextual divergence, missing entities, and added entities.

\clearpage

DPO is represented using this equation: \[
\mathcal{L}_{\text{DPO}} = -\mathbb{E}_{(x, y_+, y_-)} \left[ \log \sigma \left( r_\theta(y_+, x) - r_\theta(y_-, x) \right) \right]
\]

where, $x$: The input prompt or context, $y_+$: The preferred output (response chosen by human or other evaluators), $y_-$: The less preferred output, $r_\theta(y, x)$: The log-likelihood ratio or scoring function, typically computed by the model to reflect its belief about the relative quality of an output $y$
 given the input $x$, $\sigma(z) = \frac{1}{1 + e^{-z}}$: The sigmoid function, $\log \sigma \left( r_\theta(y_+, x) - r_\theta(y_-, x) \right)$: Represents the contrastive loss for a single pair of outputs.

 DPO essentially try to minimize log error between a pair of text, here we want to minimize ECD. Therefore we introduce a new setup of DPO, we call it RADIANT: \[
\mathcal{L}_{\text{RADIANT}} = -\mathbb{E}_{(x, y_+, y_-)} \left[ \log \sigma ( \{\frac{1}{m}\sum_{i=0}^{m} COMMON-NE_{m}^{JSD}\} + \sum_{i=0}^{n}\frac{i}{n}*\sigma_{CNE}^{JSD} - \sum_{i=0}^{n}\frac{i}{n}*\sigma_{CNE}^{JSD} ) \right]
\]


\section{DPO-based Entity-Context Divergence}

DPO aims to align AI-generated content with human preferences by optimizing for a preference-aware loss function. When applied to Entity-Context Divergence (ECD), DPO not only ensures that the AI system generates content that aligns with a retrieved context, but also that the generated content is preferred over alternative generations. This section outlines how ECD can be incorporated into DPO using preference-aware objectives.

\subsection{ECD Calculation}
The calculation of ECD for both $C_g^+$ and $C_g^-$ follows the BoW-based approach. The ECD for a given context $C_g$ relative to a retrieved context $C_r$ is computed as:
\begin{equation}
    ECD(C_r, C_g) = \frac{1}{|E_r \cap E_g|} \sum_{e \in E_r \cap E_g} \left( 1 - \frac{|W_r(e) \cap W_g(e)|}{|W_r(e) \cup W_g(e)|} \right) + \frac{\sum_{e \in E_{\text{missing}}} \text{rank}(e) \cdot \sigma}{n_{\text{common}}} + \frac{\sum_{e \in E_{\text{added}}} \text{rank}(e) \cdot \sigma}{n_{\text{common}}}
\end{equation}
where $W_r(e)$ and $W_g(e)$ denote the sets of words surrounding entity $e$ in $C_r$ and $C_g$, respectively.

To define the DPO-based Entity-Context Divergence (ECD), we introduce the following notations:
\begin{itemize}
    \item $C_r$: The retrieved context.
    \item $C_g^+$: The preferred AI-generated context.
    \item $C_g^-$: The non-preferred AI-generated context.
    \item $E_r$: The set of unique entities in the retrieved context $C_r$.
    \item $E_g^+$: The set of unique entities in the preferred context $C_g^+$.
    \item $E_g^-$: The set of unique entities in the non-preferred context $C_g^-$.
    \item $\pi(y \mid x)$: The policy of selecting output $y$ given input $x$.
    \item $s^+$: The ECD score for the preferred context $C_g^+$.
    \item $s^-$: The ECD score for the non-preferred context $C_g^-$.
    \item $\gamma$: A hyperparameter controlling the trade-off between likelihood and ECD score difference.
\end{itemize}

\subsection{DPO Statistical Loss with ECD}
The DPO objective aims to ensure that the AI system generates content that aligns with human preferences. This objective is traditionally formulated as a log-likelihood ratio of preferred vs. nonpreferred output, combined with an alignment score. By integrating ECD into this framework, we obtain a combined preference-aware loss that includes statistical and alignment objectives.

The statistical DPO loss with ECD is defined as:
\begin{equation}
    \mathcal{L}_{\text{DPO-ECD}} = - \mathbb{E}_{(x, y^+, y^-)} \left[ \log \frac{\pi(y^+ \mid x)}{\pi(y^- \mid x)} + \gamma \left( ECD(C_r, C_g^-) - ECD(C_r, C_g^+) \right) \right]
\end{equation}
where:
\begin{itemize}
    \item $\log \frac{\pi(y^+ \mid x)}{\pi(y^- \mid x)}$: Represents the log-likelihood ratio of selecting the preferred context $y^+$ over the non-preferred context $y^-$, as prescribed by the policy $\pi$.
    \item $ECD(C_r, C_g^+)$: The ECD score for the preferred context $C_g^+$ relative to the retrieved context $C_r$.
    \item $ECD(C_r, C_g^-)$: The ECD score for the non-preferred context $C_g^-$ relative to the retrieved context $C_r$.
    \item $\gamma$: A hyperparameter that controls the trade-off between preference-based loss and ECD alignment.
\end{itemize}

\paragraph{Objective Breakdown}
\begin{itemize}
    \item \textbf{Maximization over Policy ($\pi$)}: The objective seeks to find a policy that maximizes the preference for $y^+$ over $y^-$ while ensuring that $y^+$ is better aligned with the retrieved context than $y^-$.
    \item \textbf{Expectation Over Samples}: The expectation $\mathbb{E}_{(x, y^+, y^-)}$ is taken over the distribution of input-output pairs, meaning this objective is applied to a batch of training samples.
    \item \textbf{Hyperparameter $\gamma$}: This parameter controls the relative importance of the ECD alignment compared to the statistical preference loss.
\end{itemize}

\subsection{Explanation of the Components}
\begin{itemize}
    \item \textbf{Statistical Preference Loss}: The term $\log \frac{\pi(y^+ \mid x)}{\pi(y^- \mid x)}$ ensures that the model prefers the output $y^+$ over $y^-$ according to the policy $\pi$.
    \item \textbf{ECD Alignment Loss}: The term $\gamma \left( ECD(C_r, C_g^-) - ECD(C_r, C_g^+) \right)$ ensures that the context $y^+$ has a better ECD alignment score relative to the retrieved context $C_r$ than $y^-$.
    \item \textbf{Trade-off Parameter}: The parameter $\gamma$ balances the influence of the statistical preference loss and the ECD alignment loss.
\end{itemize}

This formulation ensures that the AI-generated context $y^+$ is not only aligned with human preferences but also contextually consistent with the retrieved context $C_r$. By jointly optimizing these two objectives, the resulting AI-generated content is more faithful, contextually relevant, and user-aligned.

\subsection{Gradient Calculation of DPO-ECD Loss}
The calculation of gradients for the DPO-ECD loss is crucial for optimizing the policy $\pi$. The gradient of the DPO-ECD loss with respect to the policy parameters $\theta$ is given by:
\begin{equation}
    \nabla_{\theta} \mathcal{L}_{\text{DPO-ECD}} = - \mathbb{E}_{(x, y^+, y^-)} \left[ \nabla_{\theta} \log \frac{\pi(y^+ \mid x)}{\pi(y^- \mid x)} + \gamma \nabla_{\theta} \left( ECD(C_r, C_g^-) - ECD(C_r, C_g^+) \right) \right]
\end{equation}

\paragraph{Breakdown of the Gradient Components}
\begin{itemize}
    \item \textbf{Gradient of the Log-Likelihood Ratio}: The gradient of the statistical preference loss is given by:
    \begin{equation}
        \nabla_{\theta} \log \frac{\pi(y^+ \mid x)}{\pi(y^- \mid x)} = \nabla_{\theta} \log \pi(y^+ \mid x) - \nabla_{\theta} \log \pi(y^- \mid x)
    \end{equation}
    This component encourages the policy $\pi$ to increase the likelihood of the preferred output $y^+$ relative to the non-preferred output $y^-$.

    \item \textbf{Gradient of the ECD Loss}: The gradient of the ECD alignment loss is given by:
    \begin{equation}
        \nabla_{\theta} \left( ECD(C_r, C_g^-) - ECD(C_r, C_g^+) \right) = \nabla_{\theta} ECD(C_r, C_g^-) - \nabla_{\theta} ECD(C_r, C_g^+)
    \end{equation}
    Each ECD component is further differentiated with respect to the parameters of $\pi$, considering both shared entity divergence, missing entity penalties, and added entity penalties.
\end{itemize}

\paragraph{Optimization Strategy}
The gradient can be used to update the policy parameters $\theta$ using gradient descent or other optimization methods. The total gradient is the sum of the preference-based gradient and the alignment-based gradient, both influence the policy updates to favor content that aligns with human preferences and maintains contextual alignment with $C_r$.